\documentclass{article}
\usepackage{amsfonts}
\usepackage{amsmath}
\usepackage{amssymb}
\usepackage{epsfig}
\usepackage{url}
\usepackage{algorithmic,algorithm}
\usepackage{subfigure}

\begin{document}

\title{Learning in the Model Space for Fault Diagnosis}
\author{Huanhuan~Chen, Peter~Ti\v{n}o, Xin~Yao, and Ali Rodan \thanks{
The authors are with The Centre of Excellence for Research in Computational
Intelligence and Applications (CERCIA), School of Computer Science,
University of Birmingham, Birmingham B15 2TT, United Kingdom, email:
\{H.Chen, P.Tino, X.Yao, A.A.Rodan\}@cs.bham.ac.uk.}}

\date{}
\maketitle

\begin{abstract}
The emergence of large scaled sensor networks facilitates the collection of
large amounts of real-time data to monitor and control complex engineering
systems. However, in many cases the collected data may be incomplete or
inconsistent, while the underlying environment may be time-varying or
un-formulated. In this paper, we have developed an innovative cognitive
fault diagnosis framework that tackles the above challenges. This framework
investigates fault diagnosis in the model space instead of in the signal
space. Learning in the model space is implemented by fitting a series of
models using a series of signal segments selected with a rolling window. By
investigating the learning techniques in the fitted model space, faulty
models can be discriminated from healthy models using one-class learning
algorithm. The framework enables us to construct fault library when unknown
faults occur, which can be regarded as cognitive fault isolation. This paper
also theoretically investigates how to measure the pairwise distance between
two models in the model space and incorporates the model distance into the
learning algorithm in the model space. The results on three benchmark
applications and one simulated model for the Barcelona water distribution
network have confirmed the effectiveness of the proposed framework.
\end{abstract}


\section{Introduction}

\label{introducation}

The smooth operation of complex engineering systems is crucial to the modern
society. To ensure reliability, safety and availability of such complex
systems, large amounts of real-time data will be collected to detect and
diagnose faults as soon as possible. Therefore designing an intelligent
real-time system for fault diagnosis has been receiving considerable
attention both from industry and academia.

The fault diagnosis procedure can be investigated in the following three
steps: (i) fault detection is to determine whether a fault has occurred or
not; (ii) fault isolation aims to determine the type/location of fault; and
(iii) fault identification estimates the magnitude or severity of the fault.
In some cases, the issues of fault isolation and fault identification are
interwoven, since they both determine the type of fault that has occurred.

In recent years, there has been a lot of research in the design and analysis
of fault diagnosis schemes for different dynamic systems (for example, \cite%
{Chen99,Gertler98}). A significant part of the research has focused on
linear dynamical systems, where it is possible to obtain rigorous
theoretical results. More recently, considerable effort has been devoted to
the development of fault diagnosis schemes for nonlinear systems with
various kinds of assumptions and fault scenarios \cite%
{Zhang02,zhang2005sensor,yan2007nonlinear}.

These traditional fault diagnosis approaches rely, to a large degree, on the
mathematical model of the \textquotedblleft normal\textquotedblright\
system. If such a mathematical model is available, then fault diagnosis is
achieved by comparing actual observations with the prediction of the model.
Most autonomous fault diagnosis algorithms are based on this methodology.
However, for complex engineering systems operating in unformulated or
time-varying environments, such mathematical models may not be accurate or
even unavailable at all. Therefore, it is necessary to develop cognitive
fault diagnosis methods mainly based on the collected real-time data.

In this contribution we present a novel framework for dealing with fault
detection to fault isolation if no, or very limited knowledge is provided
about the underlying system. We do not assume that we know the type, the
number or the functional form of the faults in advance. The core idea is to
transform the signal into a higher dimensional \textquotedblleft dynamical
feature space\textquotedblright\ via reservoir computation models and then
represent varying aspects of the signal through variation in the linear
readout models trained in such dynamical feature spaces. In this way parts
of the signal captured in a rolling window will be represented by the
reservoir model with the readout mapping fitted in that window.

Dynamic reservoirs of reservoir models have been shown to be `generic' in
the sense that they are able to represent a wide variety of dynamical
features of the input driven signals, so that given a task at hand only the
linear readout on top of reservoir needs to be retrained \cite%
{Lukosevicius09}. Hence in our formulation, the underlying dynamic reservoir
will be the \emph{same} throughout the signal - the differences in the
signal characteristics at different times will be captured solely by the
linear readout models and will be quantified in the function space of
readout models.

We assume that for some sufficiently long initial period the system is in a
`normal/healthy' regime so that when a fault occurs the readout models
characterizing the fault will be sufficiently `distinct' from the normal
ones. A variety of novelty/anomaly detection techniques can be used for the
purposes of detection of deviations from the `normal'. In this contribution
we will use one-class support vector machines (OCS) \cite{Scholkopf01}
methodology in the readout model space. As new faults occur in time they
will be captured by our incremental fault library building algorithm
operating in the readout model space.

There have been other learning based approaches on fault detection and
diagnosis, e.g. \cite{Vemuri97,Palade10,Venkatasubramanian03,Kankar11}. For
example, in \cite{Venkatasubramanian03}, when neural network is expanded or
the topology of the network is changed to accommodate new faults or
unexpected dynamics, the network should be retrained \cite%
{Venkatasubramanian03}. Later on, Barakat et al. proposed to use self
adaptive growing neural network for faults diagnosis \cite{Barakat01}. They
applied wavelet decomposition and used the variance and Kurtosis of the
decomposed signals as features. In 2009, Y{\'{e}}lamos et. al \cite%
{Yelamos09} proposed to use support vector machines for fault diagnosis in
chemical plants. Crucially, most of the current learning based approaches
are formulated in the supervised learning framework, assuming that all fault
patterns are known in advance. This can clearly be unrealistic.

The contributions of this paper are as follows: a) we propose a novel
learning framework for cognitive fault diagnosis; b) the framework is based
on learning in the model space (as opposed to the traditional data space) of
readout models operating on the dynamic reservoir feature space representing
parts of signals; c) we propose to use incremental one class learning in the
readout model space for fault detection/isolation and dynamic fault library
building.

The rest of this paper is organized as follows. Section \ref{model_space}
introduces deterministic reservoir computing and the framework of
\textquotedblleft learning in the model space\textquotedblright , followed
by the incremental one class learning algorithm for cognitive fault
diagnosis in Section \ref{Cognitive}. The experimental results and analysis
are reported in Section \ref{experiment}. Finally, Section \ref{conclusion}
concludes the paper and presents some future work.

\section{Deterministic Reservoir Computing and Learning in the Model Space}

\label{model_space}

This section introduces deterministic reservoir model to fit multiple-input
and multiple-output (MIMO) signals. Then, we introduce the framework of
\textquotedblleft learning in the model space\textquotedblright\ for fault
diagnosis.

\subsection{Deterministic Reservoir Computing}

Reservoir Computing (RC) \cite{Lukosevicius09} is a recent class of state
space models based on a \textquotedblleft fixed\textquotedblright\ randomly
constructed state transition mapping, realized through so-called reservoir
and an trainable (usually linear) readout mapping from the reservoir.
Popular RC methods include Echo State Networks (ESNs) \cite{Jaeger01},
Liquid State Machines \cite{Maass02} and the back-propagation decorrelation
neural network \cite{Steil04}.

In this paper, we will focus on Echo State Networks. ESNs are one of the
simplest yet effective forms of RC. Generally speaking, ESNs are recurrent
neural networks with a non-trainable sparse recurrent part (reservoir) and a
simple linear readout. Typically, the reservoir connection weights as well
as the input weights are randomly generated, subjected to the
\textquotedblleft Echo State Property\textquotedblright\ \cite{Jaeger01}.

The traditional randomized RC is largely driven by a series of randomized
model building stages, which could be unstable and hard to understand,
especially for fault diagnosis. In this paper, we propose to use the
deterministic reservoir algorithm, i.e. simple cycle topology with regular
jumps (CRJ) \cite{Rodan12}, to fit the signals for fault diagnosis, since
CRJ can approach any non-linear mapping with arbitrary accuracy. Due to the
linear training, the CRJ model can be trained fast and run in real-time.

\subsection{Learning in the Model Space}

Recently, there is a new trend in the machine learning community to
represent `local' data collections through models that capture what we think
is important in the data and do machine learning on those models - this can
have benefit of more robust and more targeted learning on diverse data
collections \cite{Brodersen11}.

The idea of learning in the model space is to use models fitted on parts of
data as more stable and parsimonious representations of the data. Learning
is then performed directly in the model space, instead of the original data
space. Some aspects of the idea of learning in the model space have occurred
in different forms in the machine learning community. For example, using
generative kernels for classification (e.g. P-kernel \cite{Haussler99} or
Fisher kernel \cite{Jaakkola99}) can be viewed as a form of learning in a
model-induced feature space (see e.g. \cite{Jebara04,Bosch08}). Recently,
Brodersen et al. \cite{Brodersen11} used a generative model of brain imaging
data to represent fMRI measurements of different subjects to build a
SVM-type learner to classify these subjects into aphasic patients or healthy
controls.

In this paper, we use \textquotedblleft learning in the model
space\textquotedblright\ approach to represent chunks of signals by dynamic
models (reservoirs models with linear readout) and perform learning in the
models space of readouts. The framework is illustrated in Figure \ref%
{learninginModel}.

\begin{figure}[t]
\centering
\includegraphics[width=3in]{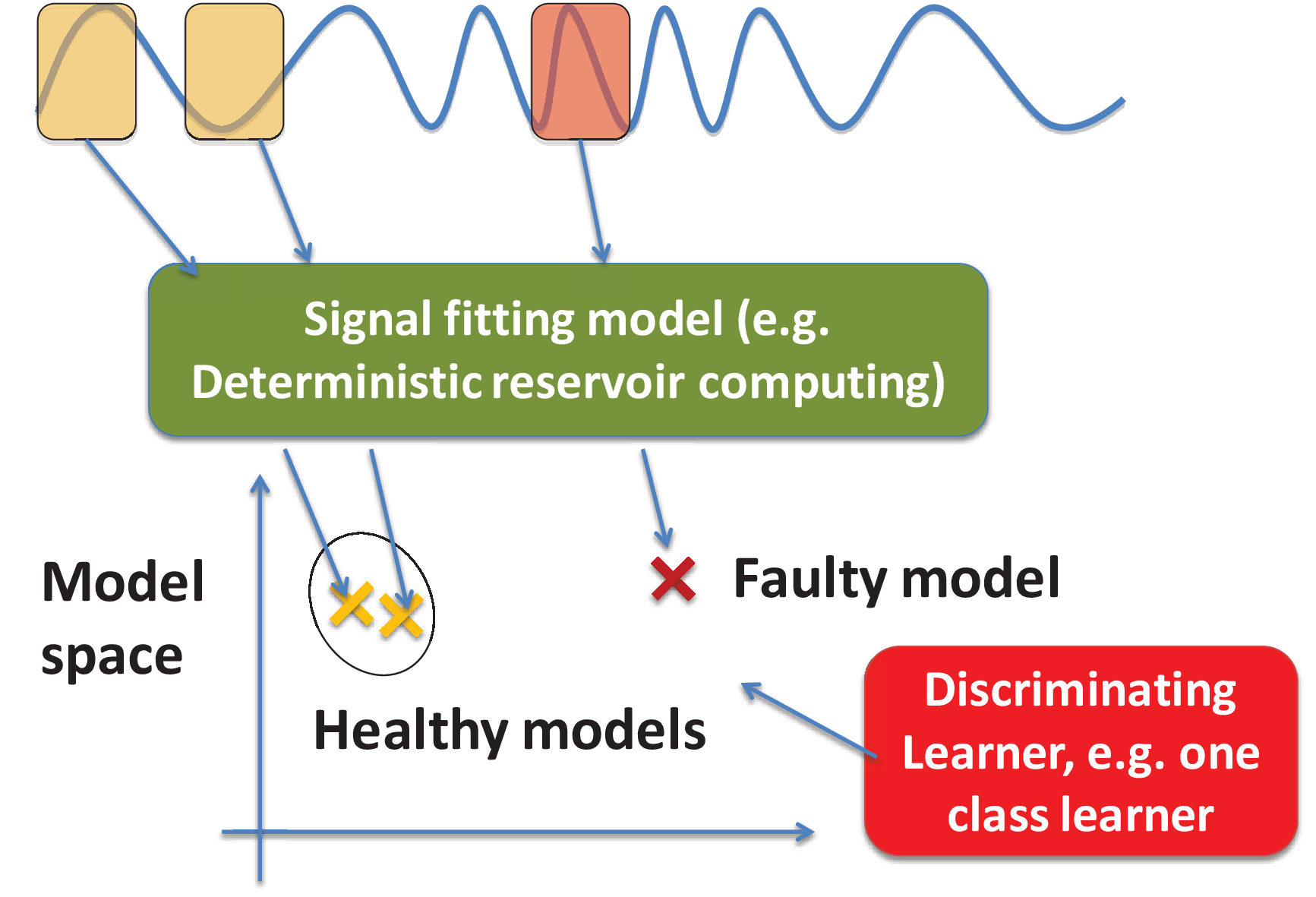}
\caption{Illustration of \textquotedblleft learning in the model
space\textquotedblright\ framework. The first stage is to fit models using
the input-output signal, i.e. generate individual points in the model space.
The second stage is to discriminate the faulty models from healthy models
using discriminating learners.}
\label{learninginModel}
\end{figure}

\subsubsection{Distance in the Model Space}

There are several ways to generate the model space from the original signal
space. One possible way is to identify parameterized models with their
parameter vectors and work in the parameter space. This, however, will make
the learning highly dependent on the particular model parameterization used.
A more satisfying approach is to use parameterization-free notions of
distance or similarities between the models.

In the model space, the $m$-norm distance between models $f_{1}(\mathbf{x})$
and $f_{2}(\mathbf{x})$ ($f_{1}$, $f_{2}:\Re ^{N}\rightarrow \Re ^{O}$) is
defined as follows:
\begin{equation*}
L_{m}(f_{1},f_{2})=\left( \int_{C}D_{m}\left( f_{1}(\mathbf{x}),f_{2}(%
\mathbf{x})\right) d\mu (\mathbf{x})\right) ^{1/m}\mathbf{,}
\end{equation*}%
where $D_{m}\left( f_{1}(\mathbf{x}),f_{2}(\mathbf{x})\right) =\left\Vert
f_{1}(\mathbf{x})-f_{2}(\mathbf{x})\right\Vert ^{m}$ is a function to
measure the difference between $f_{1}(\mathbf{x})$ and $f_{2}(\mathbf{x})$, $%
\mu (x)$ is the probability density function of the input domain $\mathbf{x}$%
, and $C$ is the integral range. In this paper, we adopt $m=2$ and first
assume that $x$ is uniformly distributed. Of course, non-uniform $\mu (%
\mathbf{x})$ can be adopted either by using samples generated from it or by
estimating it directly using e.g. Gaussian mixture models.

In the following, we demonstrate the application of the distance definition
in the model space for linear readout models. The readout model can be
represented by the following equation%
\begin{equation*}
f(\mathbf{x})=W\mathbf{x}+\mathbf{a},
\end{equation*}%
where $\mathbf{x}=[x_{1},\cdots ,x_{N}]^{T}$ is a state vector or basis
function, $N$ is the number of input variables in the model, $W$ is the
parameters ($O\times N$ matrix) in the model, $O$ is the output
dimensionality, and $\mathbf{a}=[a_{1},\cdots ,a_{o}]$ $\in \Re ^{O}$ is the
bias vector of output nodes.

Consider two readouts from the \emph{same} reservoir
\begin{eqnarray*}
f_{1}(\mathbf{x}) &=&W_{1}\mathbf{x}+\mathbf{a}_{1}, \\
f_{2}(\mathbf{x}) &=&W_{2}\mathbf{x}+\mathbf{a}_{2}.
\end{eqnarray*}%
Since the sigmoid activation function is employed in the domain of the
readout, $C$ $\in \lbrack -1,1]^{N}$. Then,
\begin{eqnarray}
&&L_{2}(f_{1},f_{2})  \notag \\
&=&\left( \int_{C}\left\Vert f_{1}(\mathbf{x})-f_{2}(\mathbf{x})\right\Vert
^{2}d\mathbf{x}\right) ^{1/2}  \notag \\
&=&\left( \int_{C}\left\Vert (W_{1}-W_{2})\mathbf{x}+(\mathbf{a}_{1}-\mathbf{%
a}_{2})\right\Vert ^{2}d\mathbf{x}\right) ^{1/2}  \notag \\
&=&\left( \int_{C}\left\Vert W\mathbf{x}\right\Vert ^{2}+2\mathbf{a}^{T}W%
\mathbf{x}+\left\Vert \mathbf{a}\right\Vert ^{2}d\mathbf{x}\right) ^{1/2}
\notag
\end{eqnarray}%
where $W=W_{1}-W_{2}$, and $\mathbf{a=a}_{1}-\mathbf{a}_{2}$.

Note that for any fixed $\mathbf{a}$ and $W$%
\begin{equation*}
\int_{C}\mathbf{a}^{T}W\mathbf{x}\ d\mathbf{x}=0,
\end{equation*}
in the integral range $C$.

Therefore,
\begin{eqnarray}
L_{2}(f_{1},f_{2}) &=&\left( \int_{C}\left\Vert W\mathbf{x}\right\Vert
^{2}+\left\Vert \mathbf{a}\right\Vert ^{2}d\mathbf{x}\right) ^{1/2}  \notag
\\
&=&\left( \int_{C}\sum_{i=1}^{O}\left( \mathbf{w}_{i}^{T}\mathbf{x}\right)
^{2}+\left\Vert \mathbf{a}\right\Vert ^{2}d\mathbf{x}\right) ^{1/2}  \notag
\\
&=&\left( \frac{2^{N}}{3}\sum_{j=1}^{N}\sum_{i=1}^{O}w_{i,j}^{2}+2^{N}\left%
\Vert \mathbf{a}\right\Vert ^{2}\right) ^{1/2}  \label{dist2}
\end{eqnarray}%
where $\mathbf{w}_{i}^{T}$ is the $i$-th row of $W$, $w_{i,j}$ is the $(i,j)$%
-th element of $W$.

Scaling of the squared model distance ($L_{2}^{2}(f_{1},f_{2})$) by $2^{-N}$
we obtain
\begin{equation*}
\frac{1}{3}\sum_{j=1}^{N}\sum_{i=1}^{O}w_{i,j}^{2}+\left\Vert \mathbf{a}%
\right\Vert ^{2},
\end{equation*}%
which differs from the squared Euclidean distance of the readout parameters
\begin{equation*}
\sum_{j=1}^{N}\sum_{i=1}^{O}w_{i,j}^{2}+\left\Vert \mathbf{a}\right\Vert
^{2},
\end{equation*}%
by the factor $1/3$ applied to the differences in the linear part $W$ of the
affine readouts. Hence, more importance is given to the `offset' than
`orientation' of the readout mapping.

In the above, we assumed that the distribution of $\mathbf{x}$ is uniform in
the integral range $C$. As mentioned before, in case of non-uniform $\mu (%
\mathbf{x})$, we can either use samples generate from $\mu $ or estimate it
analytically using e.g. a Gaussian mixture model.

Assume we have $m$ sampled points $\mathbf{x}_{i}$, $i=1,2,...,m$ from $\mu $%
. Then
\begin{eqnarray}
&&L_{2}(f_{1},f_{2})  \notag \\
&=&\left( \int_{C}\left\Vert f_{1}(\mathbf{x})-f_{2}(\mathbf{x})\right\Vert
^{2}d\mu (\mathbf{x)}\right) ^{1/2}  \notag \\
&\approx &\left( \frac{1}{m}\sum_{i=1}^{m}\left\Vert f_{1}(\mathbf{x}%
_{i})-f_{2}(\mathbf{x}_{i})\right\Vert ^{2}\right) ^{1/2}.  \label{lmu}
\end{eqnarray}

Alternatively, Gaussian mixture model can be employed to represent $\mu$,
\begin{eqnarray*}
\mu (\mathbf{x}) &=&\sum_{i=1}^{K}\alpha _{i}\mu _{i}(\mathbf{x|\eta }_{i}%
\mathbf{,}\Sigma _{i}),\text{and} \\
\mu _{i}(\mathbf{x|\eta }_{i}\mathbf{,}\Sigma _{i}) &=&\frac{\mathbf{\exp }%
\left( -\frac{1}{2}(\mathbf{x}-\mathbf{\eta }_{i})^{T}\Sigma _{i}^{-1}(%
\mathbf{x}-\mathbf{\eta }_{i})\right) }{(2\pi )^{N/2}\left\vert \Sigma
_{i}\right\vert ^{1/2}},
\end{eqnarray*}%
where $\sum_{i=1}^{K}\alpha _{i}=1$ and $N$ is the dimensionality of $%
\mathbf{x}$.

Then, the distance $L_{2}(f_{1},f_{2})$ can be obtained as follows:
\begin{eqnarray}
&&L_{2}(f_{1},f_{2})  \notag \\
&=&\left( \int_{C}(f_{1}(\mathbf{x})-f_{2}(\mathbf{x}))^{2}d\mu (\mathbf{x}%
)\right) ^{1/2}\mathbf{,}  \label{lmu2} \\
&=&\sum_{i=1}^{K}\alpha _{i}\left\{
\begin{array}{c}
trace(W^{T}W\Sigma _{i})+\mathbf{\eta }_{i}^{T}W^{T}W\mathbf{\eta }_{i} \\
+2\mathbf{a}^{T}W\mathbf{\eta }_{i}+\mathbf{a}^{T}\mathbf{a}%
\end{array}%
\right\} .  \notag
\end{eqnarray}

\section{Incremental One Class Learning for Cognitive Fault Diagnosis}

\label{Cognitive}

\begin{algorithm}[h]
\caption{Incremental One Class Learning for Cognitive Fault
Detection} \label{cogFD}
\begin{algorithmic}[1]
\STATE{\textbf{Input:} multiple-input and multiple-output
data stream $\mathbf{s}%
_{1},\cdots ,\mathbf{s}_{t},\mathbf{s}_{t+1}\cdots
$, where $\mathbf{s}%
_{t}=(u_{1},\cdots ,u_{V},y_{1},\cdots ,y_{O})^{T}$, $%
V$ is the number of signal inputs
and $O$ is the number of outputs. The data segment $\mathbf{s}%
_{1},\cdots ,\mathbf{s}_{t}$ are normal states of the
system; parameters ($\sigma$ and $\nu$) of one-class SVMs; window size $m$.}%
\STATE{\textbf{Output:} model library $lib$.}%
\FOR{each sliding window $%
(\mathbf{s}_{i},\cdots ,\mathbf{s}_{i+m-1}),1\leq
i\leq t+1-m$}%
\STATE{Fit deterministic reservoir computing model.} %
\STATE{$drc(\mathbf{s}_{i},\cdots ,\mathbf{s}_{i+m-1})\rightarrow f_{i}$}%
\ENDFOR %
\STATE{Calculate the pairwise model distance matrix $\mathbf{L}%
_{2}(f_{i},f_{j}),1\leq i,j\leq t+1-m$ according to Equation
(\ref{dist2})} %
\STATE{Apply one class SVMs: $OCS(\mathbf{L}_{2},\sigma
,\nu )\rightarrow \Theta_{0}$ and add $\Theta_{0}$ in the model library $%
lib=\{\Theta_{0}\}$.} %
\FOR{sliding window $(\mathbf{s}_{j},\cdots,%
\mathbf{s}_{j+m-1}),j>t$}
\STATE{$drc(\mathbf{s}_{j},\cdots ,\mathbf{s}%
_{j+m-1})\rightarrow f_{j}$;} %
\IF{$f_{j}$ belongs to a known fault $\Theta_{k}$ in the $lib$}
\STATE{update $\Theta_{k}$ with $f_{j}$ and empty candidate pool;} %
\ELSE %
\STATE{put $f_{j}$ in the candidate pool;}%
\ENDIF %
\IF{size of candidate pool $>0.5*m$} %
\STATE{build a new model $\Theta_{k+1}$ with candidate pool} %
\STATE{Add $\Theta_{k+1}$ to $lib$ and empty candidate pool}%
\ENDIF%
\ENDFOR %
\end{algorithmic}
\end{algorithm}

In fault diagnosis, it should be determined whether a running
sub-system/component is in a normal operation condition, or whether a faulty
situation is occurring. It is relatively cheap and simple to obtain
measurements from a normally working system (although sampling from all
possible normal situations might still be expensive). In contrast, sampling
from faulty situations requires the system to break down in various ways to
obtain faulty measurement examples. The construction of a fault library will
therefore be very expensive, or completely impractical. In this section, we
focus on this challenge and aim to develop an algorithm that can identify
unknown faults and construct a fault library dynamically, which will
facilitate fault isolation based on this library.

Based on the \textquotedblleft learning in the model
space\textquotedblright\ framework (Figure \ref{learninginModel}), one class
learning \cite{Scholkopf01} will be employed in the model space for fault
diagnosis. One-class classification is a special type of classification
algorithm. One-class SVMs are to discover a hyperplane that has maximal
distance to the origin in the kernel feature space with the given training
examples falling beyond the hyperplane \cite{Scholkopf01}.

Note that the signal characteristics can change at different positions of
the rolling window. That means that the underlying measure $\mu $ over
reservoir activations $\mathbf{x}$ can change. Consider two readouts $f_{i}$
and $f_{j}$ obtained from two rolling window positions $i$ and $j$. If
reservoir activations in positions $i$ and $j$ are considered we would
obtain two distances $L_{\mu _{i}}(f_{i},f_{j})$ and $L_{\mu
_{j}}(f_{i},f_{j})$, respectively\footnote{%
The measures $\mu _{k}$ will be represented by reservoir activation samples
at window position $k$.}. The distance $f_{i}$, $f_{j}$ based on the
sampling approach is then
\begin{equation*}
\tilde{L}_2(f_{i},f_{j})=L_{\mu _{i}}(f_{i},f_{j})+L_{\mu _{j}}(f_{i},f_{j}).
\label{sample_distance}
\end{equation*}

In this paper, we propose an algorithm that can construct the fault library
online. The idea is to use each one-class learner to represent each
fault/sub-fault segment by using the \textquotedblleft learning in the model
space\textquotedblright\ approach. In the beginning, a normal one-class
learner $\Theta _{0}$ will be constructed based on the normal signal
segments. With the rolling window moving forward, we continually apply $%
\Theta _{0}$ to judge whether a fault occurs. If a fault is coming, we will
train a new one-class-learner $\Theta _{i}$ for fault $i$. Then, we keep
monitoring the signal and determine whether the ongoing signal segment
belongs to either normal state or a known fault. If not, a new one-class
learner $\Theta _{i}$ will be built and included in the model library. The
algorithm is illustrated in Algorithm \ref{cogFD}, which includes the
following major steps:

\begin{enumerate}
\item Normal data preparation by applying deterministic reservoir model $drc$
to the rolling windows (size $m$) in the first $t$ steps, i.e. the
\textquotedblleft normal\textquotedblright\ regime is sequentially induced.
(Lines 3-6)

\item Calculate the pairwise model distance matrix $\mathbf{L}%
_{2}(f_{i},f_{j})$ and employ one class SVMs (OCS) to obtain the normal
class $\Theta _{0}$. (Lines 7-8)

In one class SVMs, Gaussian RBF kernel is employed with the data distance
replaced by the \emph{model distance} $L_{2}(f_{i},f_{j})$;
\begin{equation*}
\phi _{\sigma }(f_{i},f_{j})=\exp \left\{ -\sigma \cdot
L_{2}(f_{i},f_{j})\right\} \text{.}
\end{equation*}

\item With the rolling window moving forward, if a new $f_{j}$ belongs to an
existing model $\Theta _{k}$\footnote{%
If the new point $f_{j}$ is classified to more than one model by one-class
SVMs, count the point in the last model because of sequential correlation.},
update the existing $\Theta _{k}$ with this new data $f_{j}$ and empty
candidate pool. Otherwise, put the \textquotedblleft
point\textquotedblright\ $f_{j}$ in the candidate pool. (Lines 9-15)

\item If the number of data points in the candidate pool exceeds half of the
window size $m$, construct a new one-class learner $\Theta _{k+1}$ and empty
the candidate pool. (Lines 16-18)
\end{enumerate}

In the above algorithm, the assumption is that the system is running
normally in the first $t$ steps. Although the window size $m$ should be
relatively large (e.g. $>300$ time steps) to accurately fit the dynamic
models (e.g. deterministic reservoir computing in this paper). The rolling
window is moved forward by one time step, which reduces fault detection
delays.

\section{Experimental Studies}

\label{experiment}

This section presents experimental results in four-\textquotedblleft
fault\textquotedblright -diagnosis scenarios, which include one synthetic
nonlinear auto-regressive moving average (NARMA) system with three different
signals, one van der Pol oscillator with three faults imposed, one benchmark
three-tank-system with three faults and Barcelona water system with 31
faults. This paper will investigate fault detectability and fault
isolationability using a number of approaches.

\subsection{Experimental Settings}

\label{settings}

\begin{table}[tbph]
\caption{Algorithms and Parameters}
\label{AlgoPara}\centering%
\resizebox {3.8in}{!}{
\begin{tabular}{|c|c|c|}
\hline
Algorithm & Space & Parameters \\ \hline
T2 & signal & - \\ \hline
DBscan & model &
\begin{tabular}{ll}
$k$ & number of neighborhood \ \ \ \  \\
$\varepsilon $ & neighborhood radius \
\end{tabular}
\\ \hline
AP-Model & model & - \\ \hline AP-Signal & signal & - \\ \hline
OCS-Model & model &
\begin{tabular}{ll}
$\sigma $ & Gaussian kernel parameter \ \  \\
$\nu $ & the upper bound of outliers \ \ \ \ \
\end{tabular}
\\ \hline
OCS-Signal & signal &
\begin{tabular}{ll}
$\sigma $ & Gaussian kernel parameter \ \ \ \  \\
$\nu $ & the upper bound of outliers \
\end{tabular}
\\ \hline
ARMAX-OCS & model &
\begin{tabular}{ll}
$\sigma $ & Gaussian kernel parameter \\
$\nu $ & the upper bound of outliers \\
$m$ & number of nodes in reservoir (25) \\
$p$ & p autoregressive terms \\
$q$ & moving average terms \\
$b$ & exogenous inputs terms\end{tabular}
\\ \hline
RC-OCS & model &
\begin{tabular}{ll}
$\sigma $ & Gaussian kernel parameter \\
$\nu $ & the upper bound of outliers \\
$m$ & number of nodes in reservoir (25)\end{tabular}
\\ \hline
DRC-OCS (sampling) & model &
\begin{tabular}{ll}
$\sigma $ & Gaussian kernel parameter \\
$\nu $ & the upper bound of outliers \\
$m$ & number of nodes in reservoir (25)\end{tabular}
\\ \hline
DRC-OCS & model &
\begin{tabular}{ll}
$\sigma $ & Gaussian kernel parameter \\
$\nu $ & the upper bound of outliers \\
$m$ & number of nodes in reservoir (25)\end{tabular}
\\ \hline
\end{tabular}}
\end{table}

In our experiments, to evaluate the \textquotedblleft learning in the model
space\textquotedblright\ framework for fault diagnosis, a number of
approaches have been adopted for comparisons. The approaches include:
Hotelling's T-squared statistic test (T2) \cite{Cho05}, a density-based
algorithm for discovering clusters in large spatial databases with noise
(DBscan) \cite{Ester96}, affinity propagation \cite{Frey07} in the model
space (AP-Model), affinity propagation in the signal space (AP-Signal), one
class SVMs \cite{Scholkopf01} in the model space (OCS-Model), one class SVMs
in the signal space (OCS-Signal), autoregressive--moving-average model with
exogenous inputs with incremental one-class leaner (ARMAX-OCS), reservoir
computing with incremental one-class leaner (RC-OCS), deterministic
reservoir computing with incremental one-class leaner (DRC-OCS) and DRC-OCS
(sampling) where the model distance matrix is estimated by sampling method
(Equations (\ref{lmu} and (\ref{sample_distance}))). Table \ref{AlgoPara}
summaries all the algorithms employed in this paper.

The signal space is generated by selecting $p$ consecutive points, i.e. $\{%
\mathbf{s}_{t},\cdots ,\mathbf{s}_{t+p-1}\}$, where $\mathbf{s}%
_{t}=(u_{1},\cdots ,u_{V},y_{1},\cdots ,y_{O})^{T}$, as a training point by
re-arranging these $p$ points to one vector. The order $p$ will be selected
in the range $[1,30]$.

In the following four data sets, we generate 3000 time steps for normal
signal and each fault signal, respectively, and employ a rolling window
(size 500) to generate a series of data segments, which are employed to
train deterministic reservoir model. In each data set, the first 1000 time
steps of the signal are normal, i.e. the first 500 models are normal with
window size 500.

The parameters of DBscan are optimized by minimizing the number of
discovered classes and the false alarm rates using the first 500 normal
points. The parameters of ARMAX are selected by minimizing the normalized
mean squared error (NMSE) in the first 1000 time steps. The parameters of
one class SVMs in OCS-Model, OCS-Signal, ARMAX-OCS, RC-OCS and DRC-OCS will
be optimized by 5-fold cross validation using the first 500 data points.

\subsection{NARMA System}

\begin{figure}[tbph]
\centering
\includegraphics[width=2.8in]{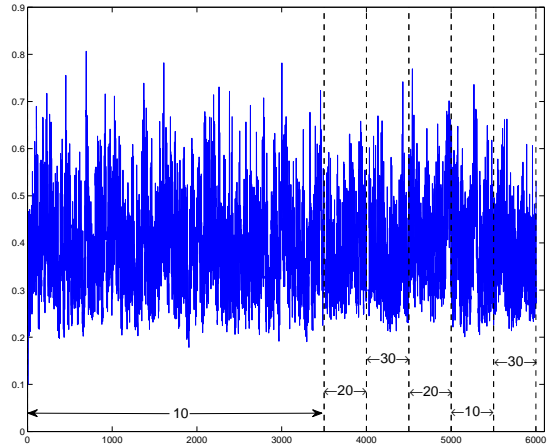}
\caption{Illustration of three NARMA sequences with different orders (10, 20
and 30).}
\label{fig:narma}
\end{figure}

\begin{figure}[btph]
\centering
\subfigure{ \label{fig:narmaPCA:a}
\includegraphics[width = 2.5in]{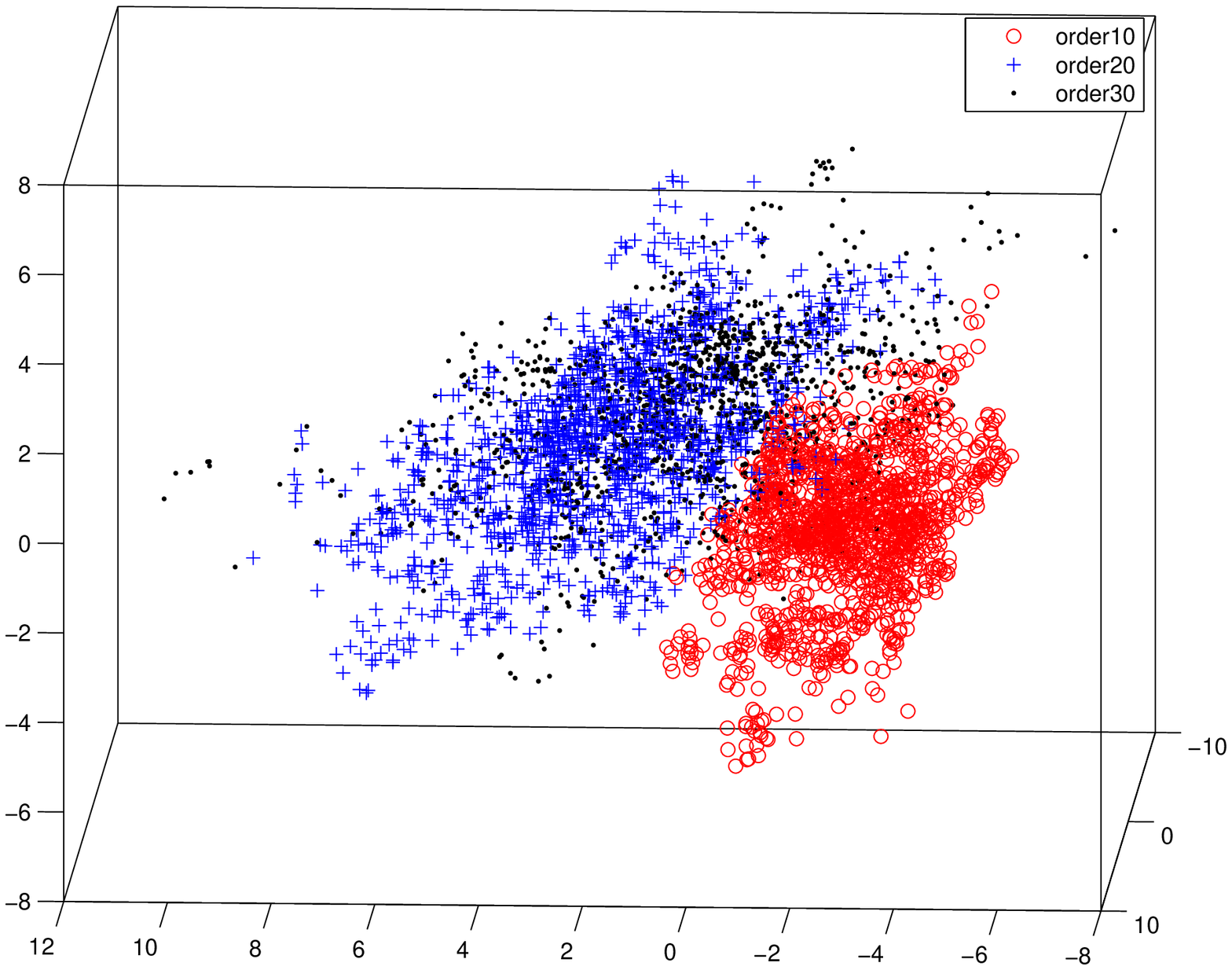}} \hspace{0.2in}
\subfigure{
    \label{fig:narmaPCA:b}    \includegraphics[width = 2.5in]{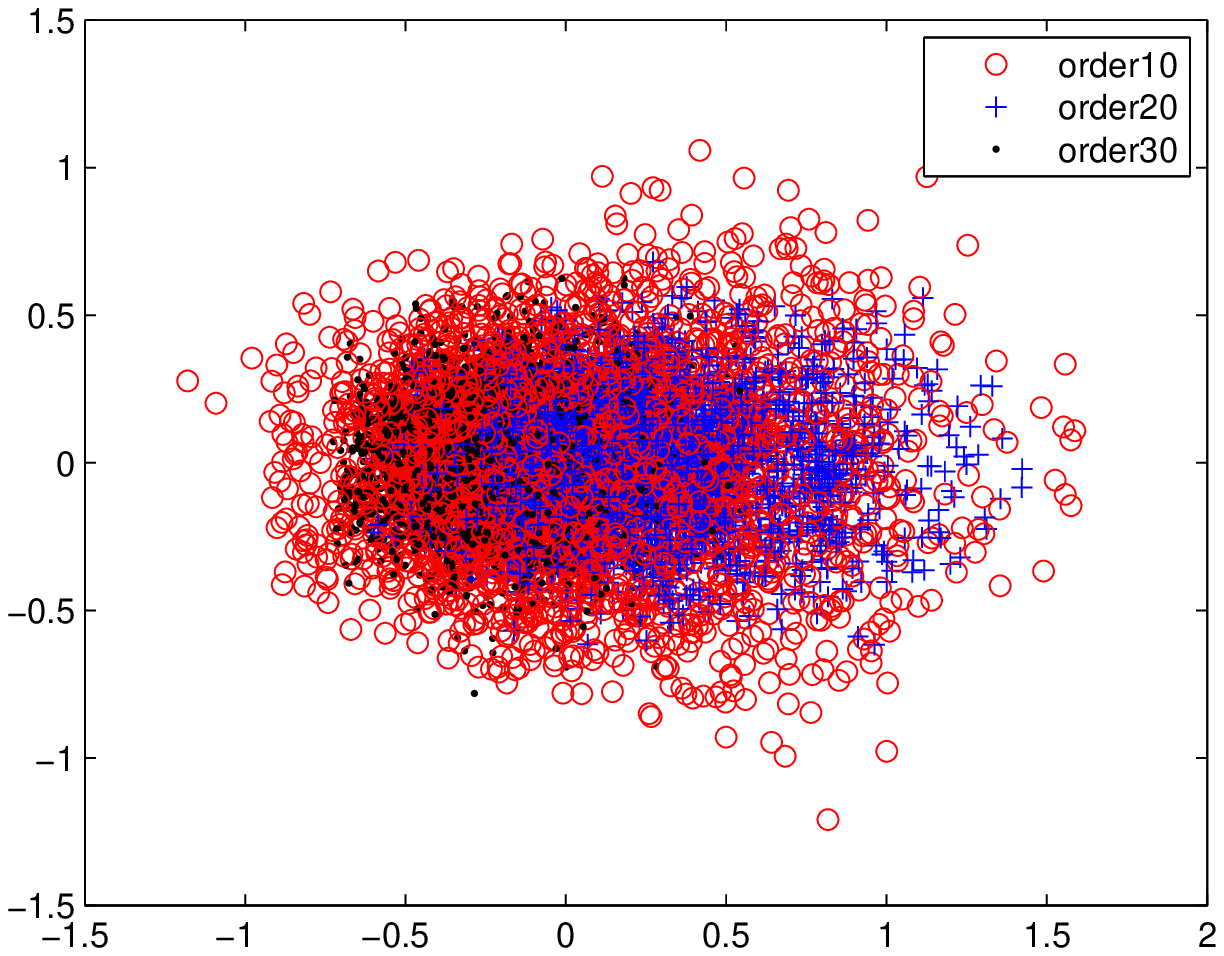}}
\caption{Visualization of the NARMA data set in the model space (top) and
signal space ($p=30$) (bottom) by multi-dimensional scaling (MDS).}
\label{fig:narmaPCA}
\end{figure}

In NARMA, the current output depends on both the input and the previous
output. Generally speaking, it is difficult to model this system due to high
non-linearity and possibly long memory. In this paper, we employed three
NARMA time series with orders $O=10,20,30$ that are given by Equations (\ref%
{eq:narma1}), (\ref{eq:narma2}) and (\ref{eq:narma3}), respectively.
\begin{equation}
y(t+1)=0.3y(t)+0.05y(t)\sum_{i=0}^{9}y(t-i)+1.5u(t-9)u(t)+0.1,
\label{eq:narma1}
\end{equation}%
\begin{equation}
y(t+1)=\tanh
(0.3y(t)+0.05y(t)\sum_{i=0}^{19}y(t-i)+1.5u(t-19)u(t)+0.01)+0.2,
\label{eq:narma2}
\end{equation}%
\begin{equation}
y(t+1)=0.2y(t)+0.004y(t)\sum_{i=0}^{29}y(t-i)+1.5u(t-29)u(t)+0.201,
\label{eq:narma3}
\end{equation}%
where $y(t)$ is the system output at time $t$, $u(t)$ is the system input at
time $t$ ($u(t)$ is an i.i.d stream generated uniformly in the interval $%
[0,0.5)$.

The three sequences are illustrated in Figure \ref{fig:narma}. The three
NARMA sequences look quite similar, and it is very difficult to separate
them based on the signal only.

Figure \ref{fig:narmaPCA} shows MDS analysis\footnote{%
Multidimensional scaling (MDS) aims to preserve the pairwise distance
between points, which is suitable to preserve the \emph{model distance} for
visualization.} of the NARMA data set in the model space (top) and in the
signal space (bottom). Based on this figure, it is relatively easier to
separate different classes in the model space, while most of the data points
overlap in the signal space. The figure confirms that the model based
representation is able to effectively represent the signals. In Table \ref%
{verse}, several supervised classification techniques have been employed to
confirm the benefits of using model space based approaches.

\subsection{Van der Pol Oscillator}

\begin{figure}[tbph]
\centering%
\subfigure{ \label{fig:vanderpol:a}
\includegraphics[width = 2.8in]{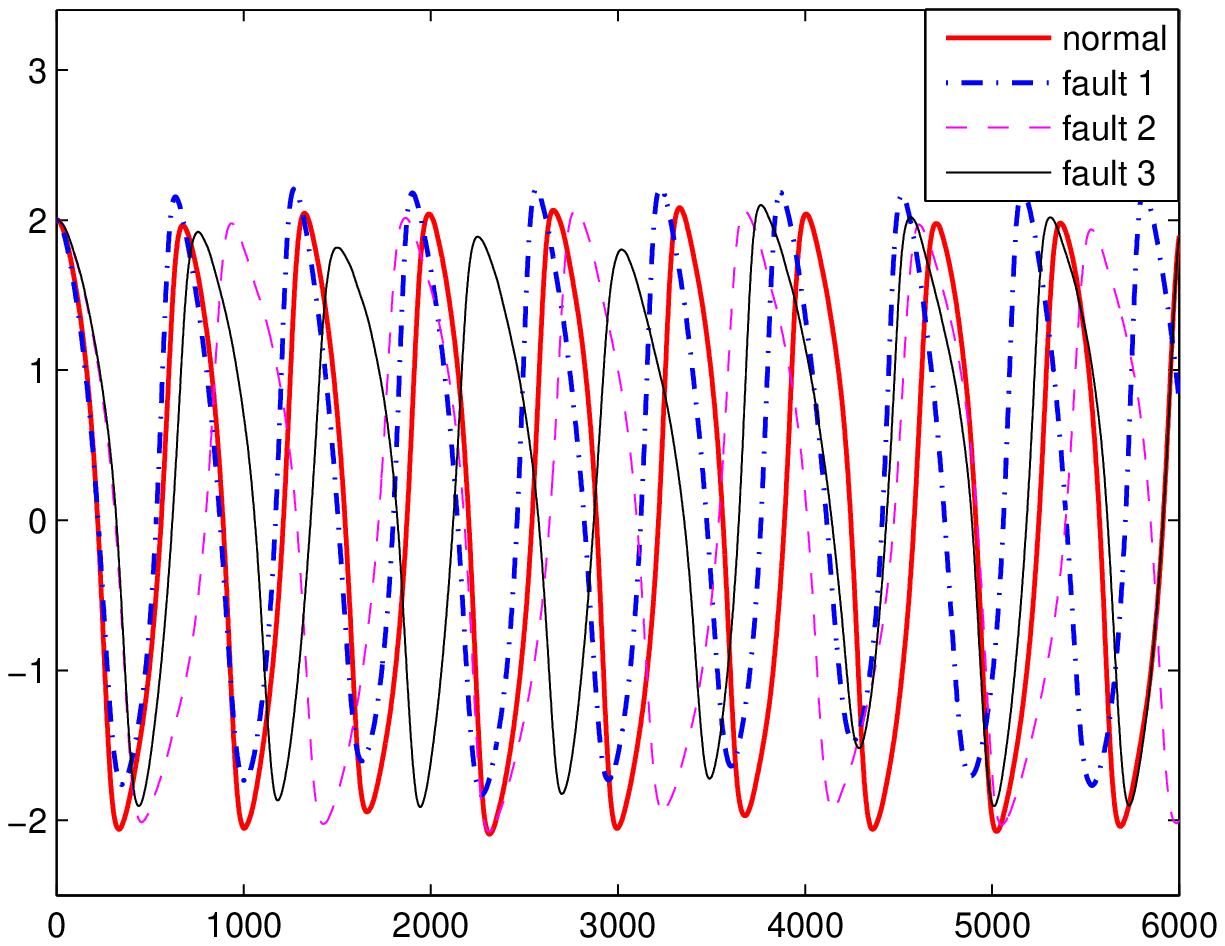}} \hspace{0.1in}
\subfigure{
    \label{fig:vanderpol:b}
    \includegraphics[width = 2.8in]{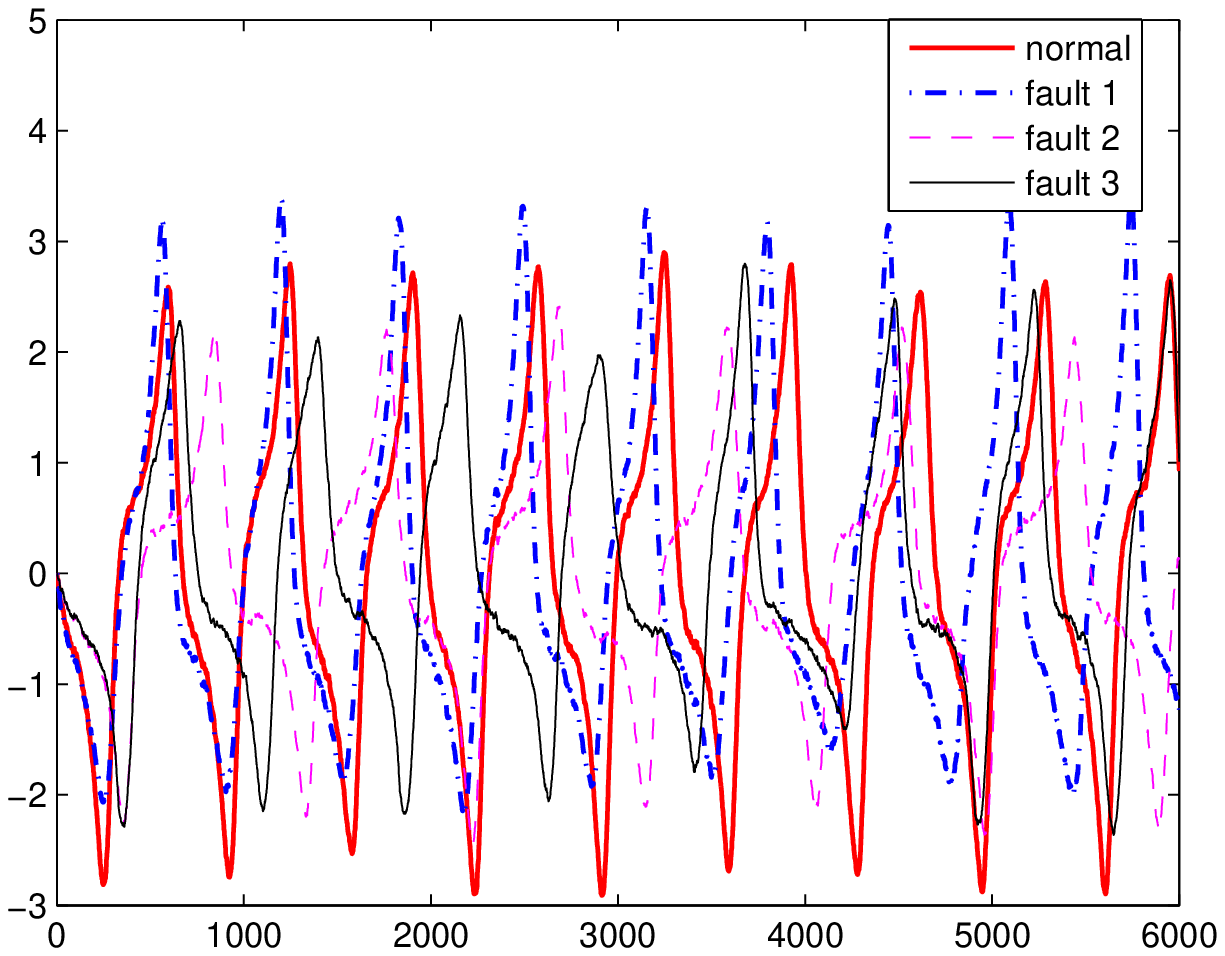}}
\caption{Illustration of Van der Pol oscillator and three different faults.
(top: $y_{1}(k)$, bottom: $y_{2}(k)$)}
\label{fig:vanderpol}
\end{figure}


A Van der Pol oscillator \cite{Kaplan95} has been a subject of extensive
research and its discrete-time expressions play an important role in the
numerical investigations. Discrete-time Van der Pol oscillator can be
obtained as follows
\begin{eqnarray*}
y_{1}(k) &=&y_{2}\Delta t+y_{1}(k-1), \\
y_{2}(k) &=&y_{2}(k-1)+y_{2}(k-1)(1-y_{1}(k-1)^{2})\Delta t \\
&&-y_{1}(k-1)\Delta t+\epsilon ,
\end{eqnarray*}%
where $\epsilon $ is Gaussian white noise with variance 0.01.

Three faults are imposed to the van der Pol oscillator by adding $0.75\sin
(y_{1}(k-1))\Delta t$, $0.75\tanh (y_{1}(k-1))\Delta t$ and $0.75\cos \left(
y_{1}(k-1)^{2}\right) $ to $y_{2}(k)$. The van der Pol oscillator and the
three faults are illustrated in Figure \ref{fig:vanderpol}.

\subsection{Three Tank System}

\begin{figure}[tbph]
\centering\includegraphics[width=2.2in]{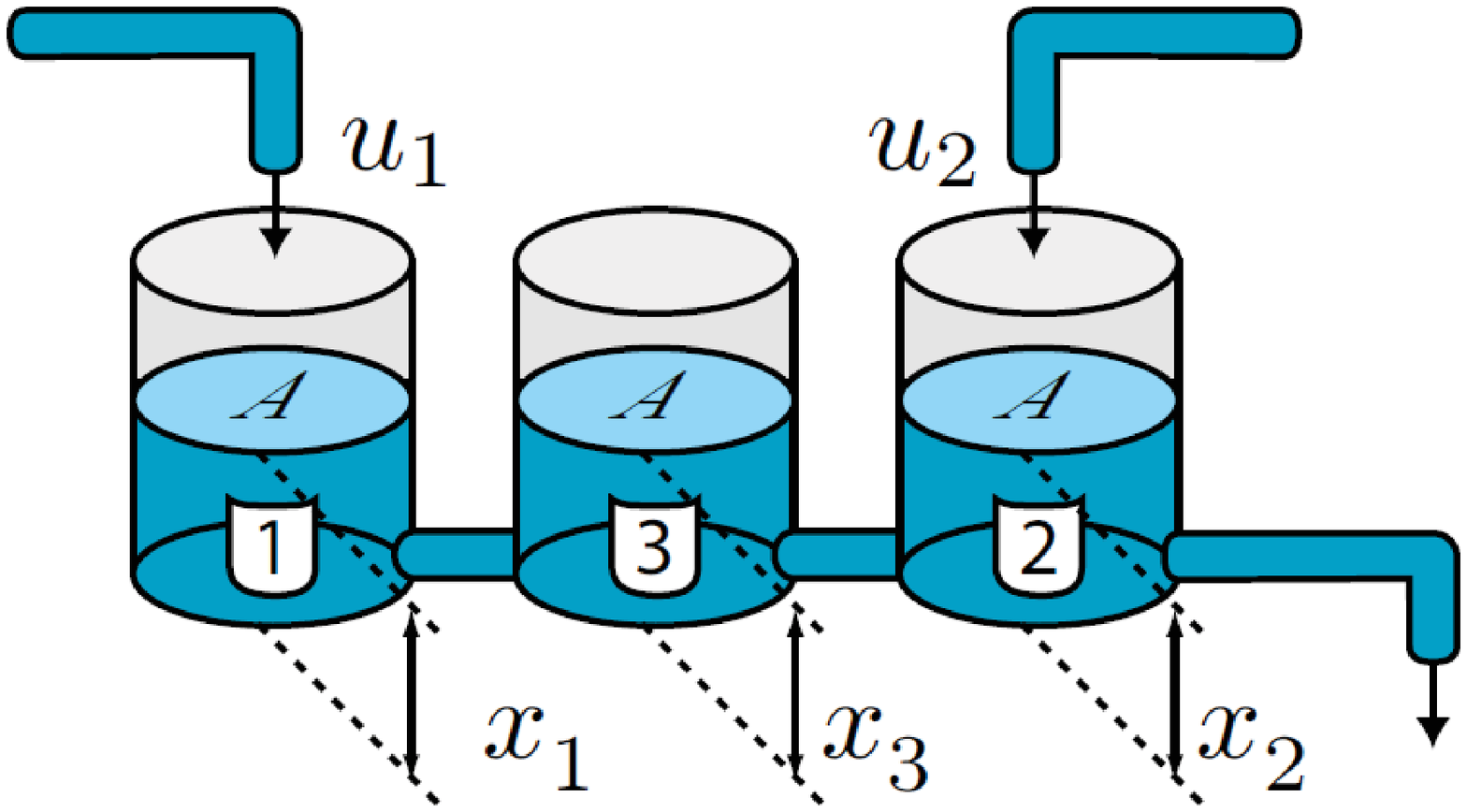}
\caption{Three tank system \protect\cite{Zhang02}.}
\label{threetank}
\end{figure}

\begin{figure*}[btph]
\centering
\subfigure{ \label{fig:threetank:a}
\includegraphics[width = 2in]{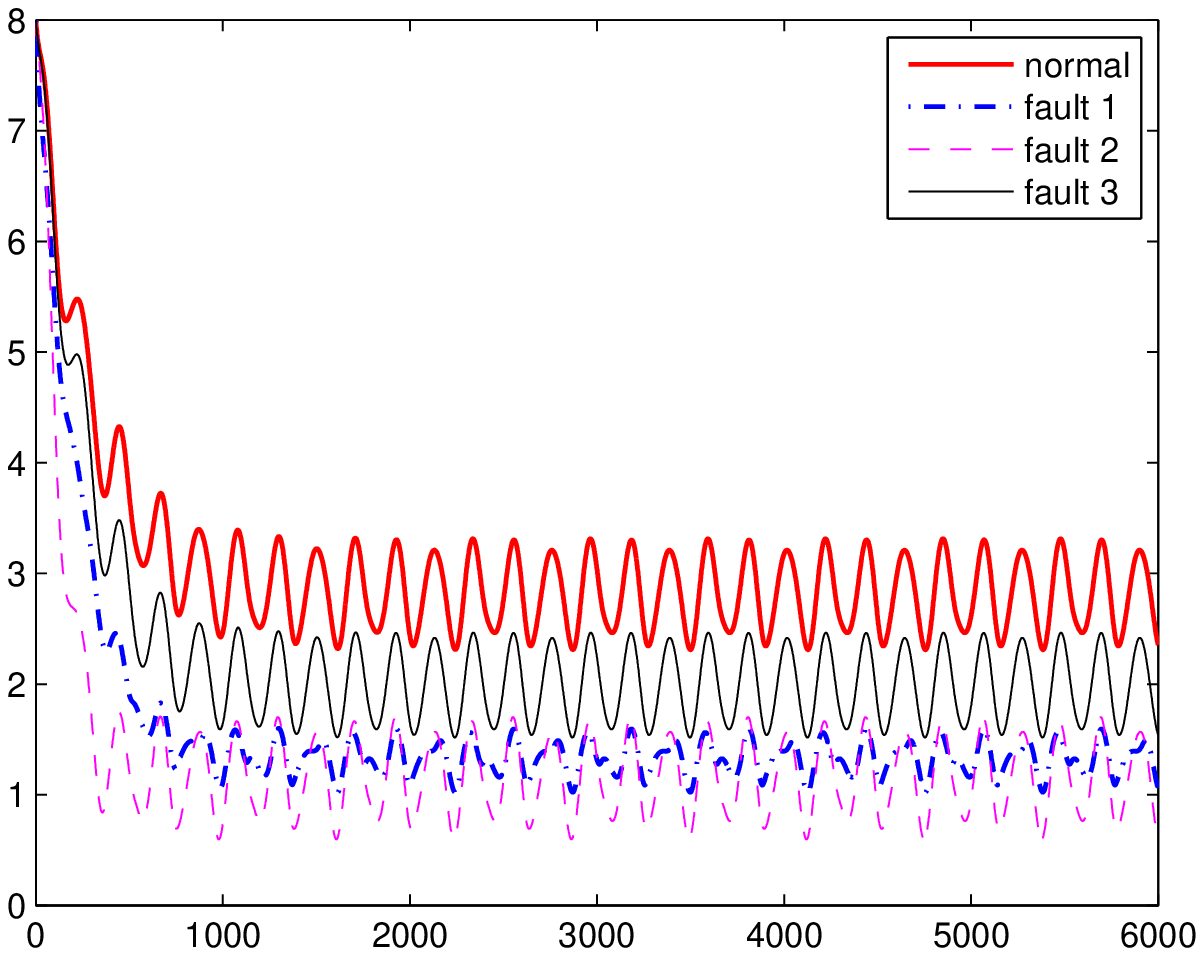}} \hspace{0.1in}
\subfigure{
    \label{fig:threetank:b}
    \includegraphics[width = 2in]{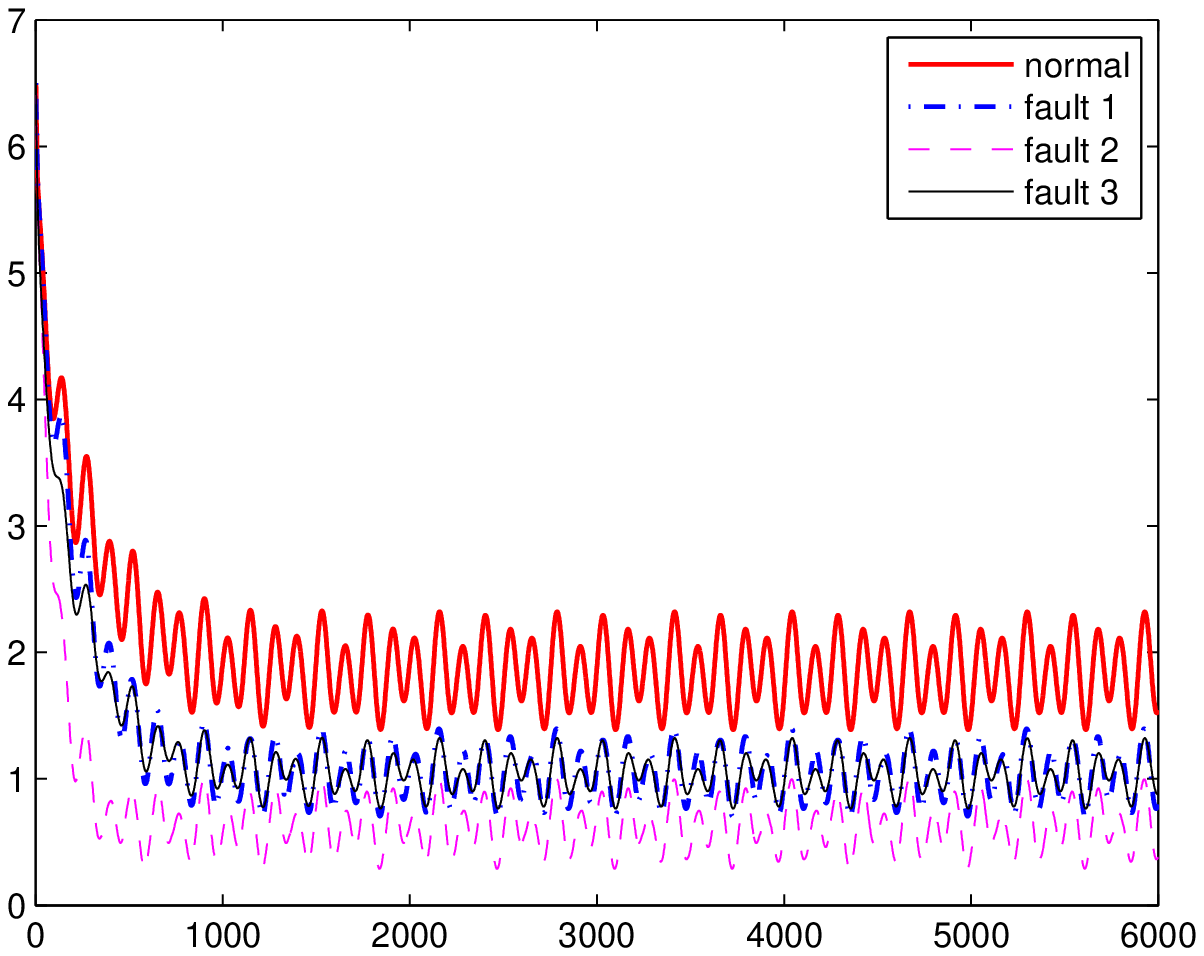}} \hspace{0.1in}
\subfigure{
    \label{fig:threetank:c}
    \includegraphics[width = 2in]{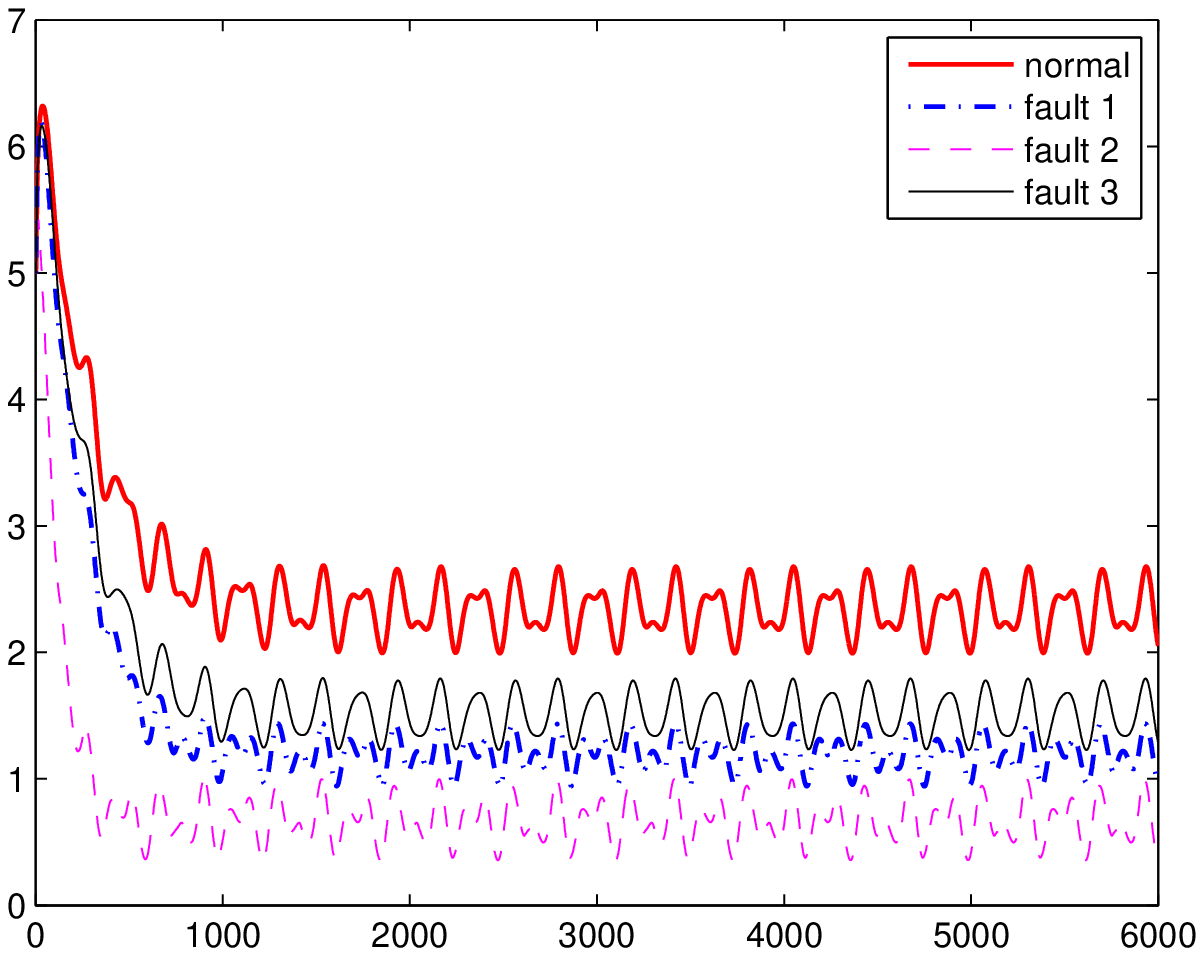}}
\caption{Illustration of levels in three tanks in the three tank system and
three different faults. (left: tank 1, middle: tank 2, right: tank 3)}
\label{fig:threetank}
\end{figure*}


A well-known three-tank problem \cite{Zhang02} in Figure \ref{threetank} is
presented to illustrate the effectiveness of the proposed algorithm. The
cross-section of these tanks is $A_{i}=1m^{2}$, and there is a cross-section
$A_{p}=0.1m^{2}$ at the end of each tank. The outflow rate is $c_{j}$, $%
i,j=1,\cdots ,3$. The level of each tank is denoted by $x_{i}$ ($0\leq
x_{i}\leq $ $10$, $i=1,\cdots ,3$).

The input flows by two pumps are denoted by $u_{i}$ with the restrictions $%
0\leq u_{i}$ $\leq $ $1m^{3}/s$, $i=1,2$. In this paper, the inflows are set
with $u_{1}(k)=0.2\cos (0.3kT_{s})+0.3$ and $u_{2}(k)=0.25\cos
(0.5kT_{s})+0.3$, respectively, and the initial levels of thanks are 8, 6.5,
and 5 meter. In the model, three faults are introduced as follows:

\begin{enumerate}
\item[1)] \textbf{Actuator fault in pump 1}: the pump is partially or fully
shutdown.

\item[2)] \textbf{Leakage in tank 3}: there is a leak circular hole with
unknown radius $0<\rho _{3}<1$ in the tank bottom.

\item[3)] \textbf{Actuator fault in pump 2}: the fault is same as fault 1
but related to pump number 2.
\end{enumerate}

Figure \ref{fig:threetank} illustrates the water levels of three tanks in
normal and three faulty situations.

\subsection{Barcelona Water Distribution Network}

\begin{figure*}[tbph]
\centering\includegraphics[angle = -90,
width=5in]{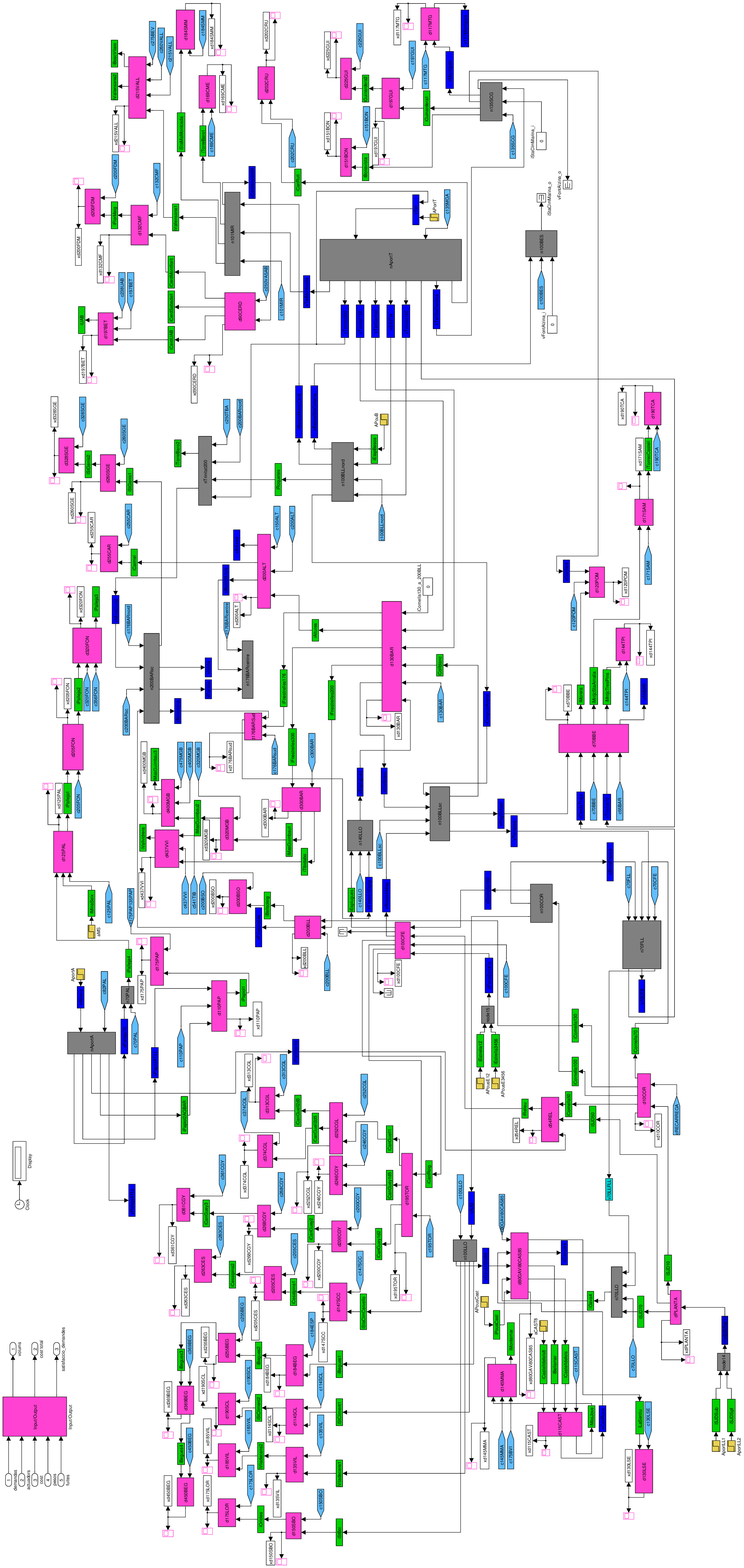}
\caption{Barcelona Water System Simulator Programmed by MATLAB Simulink
\protect\cite{Quevedo10}.}
\label{simu}
\end{figure*}

\begin{figure}[t]
\centering%
\subfigure{
    \label{fig:water:a} \includegraphics[angle = -90,width = 1.5in]{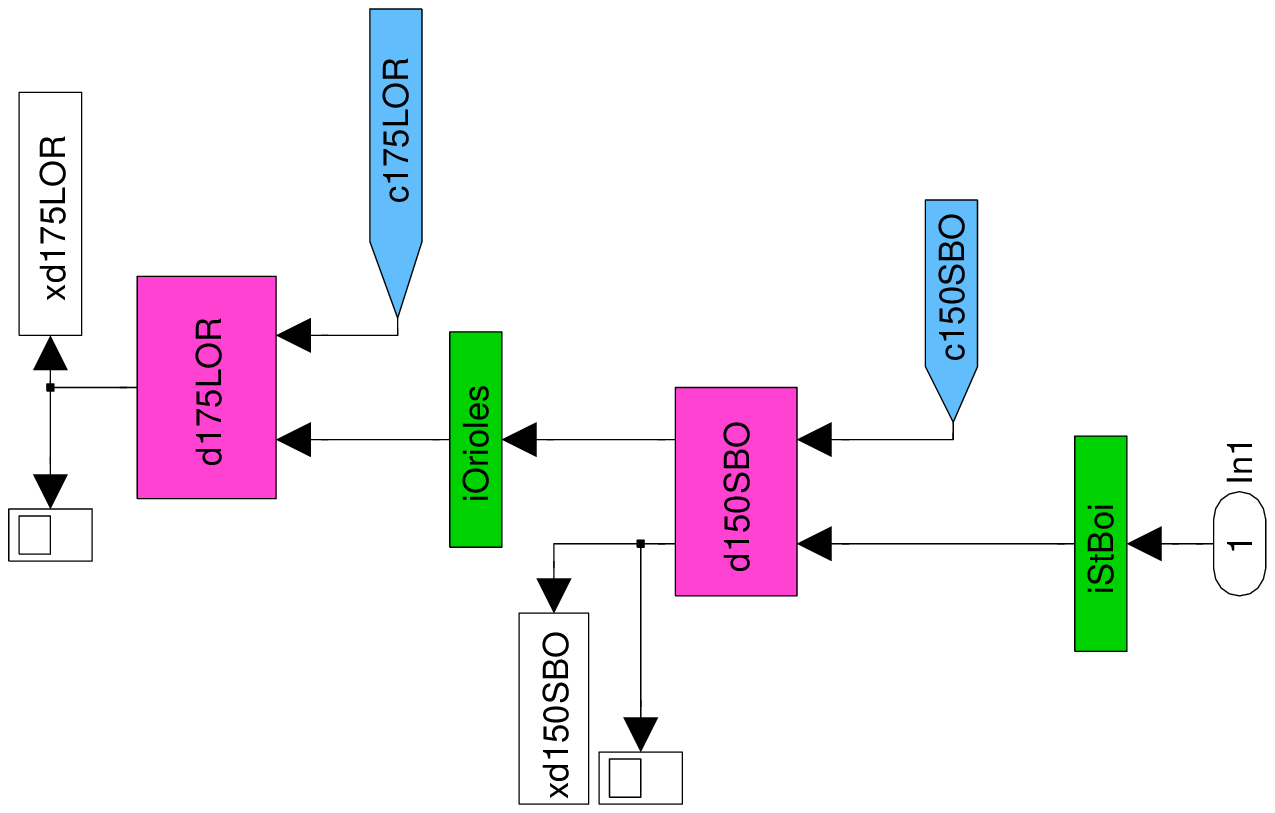}}
\hspace{0.1in}
\subfigure{
    \label{fig:water:b} \includegraphics[angle = -90,width = 1.2in]{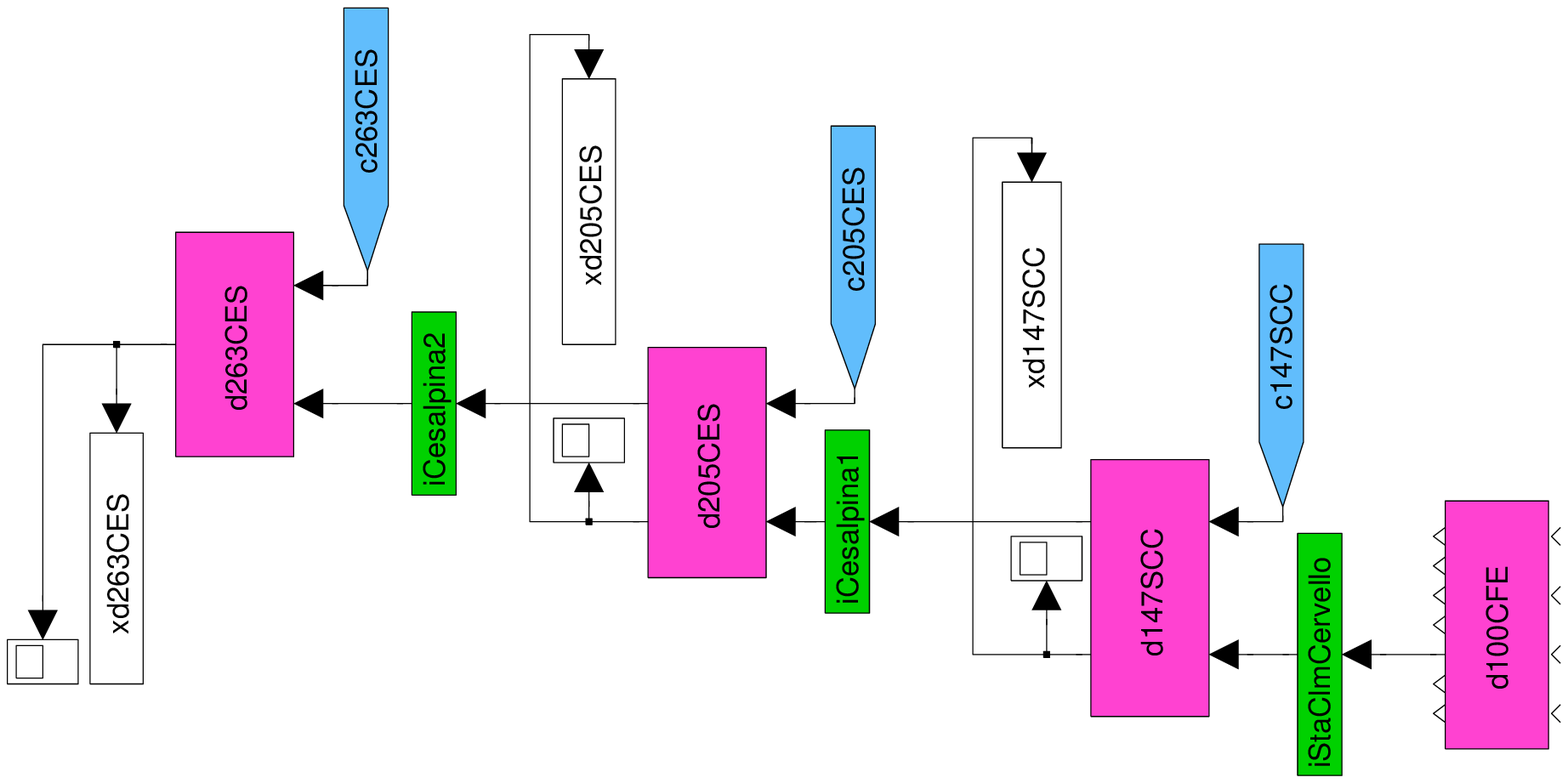}}
\caption{Subsystems of the water network where faults are introduced.
iOrioles, iStaClmCervello, iCesalpina1, iCesalpina2 are actuators
(controller). c175LOR, c147SCC, c205CES, c263CES are demand (input).
d175LOR, d147SCC, d205CES, d263CES are tank level (output).}
\label{fig:water}
\end{figure}

The next application is Barcelona Water Distribution Network (BWDN) \cite%
{Quevedo10}. BWDN supplies water to approximately 3 million consumers,
distributed in 23 municipalities in a 424 $km^{2}$ area. Water can be taken
from both surface and underground sources. From these sources, water is
supplied to 218 demand sectors through about 4645 $km$ of pipe. The complete
transport network has been modeled using 63 storage tanks, 3 surface and 6
underground sources, 79 pumps, 50 valves, 18 nodes and 88 demands.

A detailed simulation model of the BWDN has been developed using
MATLAB/Simulink \cite{Quevedo10} (Figure \ref{simu}), which has been
calibrated and validated using real data. In this simulator, we can
manipulate and inject different faults into the system. Studied faults are
introduced in the two subsystems of the network shown in Figure \ref%
{fig:water}. In the two subsystems, we introduced 31 faults, which are
detailed in Table \ref{para_fault}. These faults include actuator faults,
actuator sensor faults, demand (input) sensor faults, and tanks (output)
sensor faults. Four examples of faulty signals are illustrated in Figure \ref%
{fault}.

\begin{table}[tbph]
\caption{Parameterizations of faults. MFD stands for maximum flow/demand.}
\label{para_fault}\centering%
\resizebox {5in }{!}{
\begin{tabular}{|l|l|l|l|l|l|l|l|}
\hline ID & Faulty Element & Type & Magnitude & ID & Faulty Element
& Type & Magnitude \\ \hline 1 & iOrioles & 1 & -25\% & 17 &
iStaClmCervello & 3 & 0.01\% \\ \hline 2 & iOrioles & 2 & -25\% & 18
& iStaClmCervello & 4 & 0.5\% \\ \hline 3 & iOrioles & 2 & -10\% &
19 & iStaClmCervello & 5 & - \\ \hline 4 & iOrioles & 3 & 0.001\% &
20 & iStaClmCervello & 6 & 4 \\ \hline 5 & iOrioles & 3 & 0.1\% & 21
& iCesalpina1 & 1 & 10\% \\ \hline 6 & iOrioles & 4 & 10\% & 22 &
iCesalpina1 & 2 & -15\% \\ \hline 7 & iOrioles & 4 & 1\% & 23 &
iCesalpina1 & 3 & 0.01\% \\ \hline 8 & iOrioles & 5 & - & 24 &
iCesalpina1 & 4 & 0.75\% \\ \hline 9 & iOrioles & 6 & 2 & 25 &
iCesalpina1 & 5 & - \\ \hline 10 & c175LOR & 1 & -20\% & 26 &
iCesalpina1 & 6 & 0.75 \\ \hline 11 & c175LOR & 2 & -15\% & 27 &
c263CES & 1 & 30\% \\ \hline 12 & c175LOR & 3 & 0.01\% & 28 &
c263CES & 2 & -15\% \\ \hline 13 & c175LOR & 4 & 1\% & 29 & c263CES
& 3 & 0.025\% \\ \hline 14 & c175LOR & 5 & - & 30 & c263CES & 4 &
0.5\% \\ \hline 15 & iStaClmCervello & 1 & -15\% & 31 & c263CES & 5
& - \\ \hline 16 & iStaClmCervello & 2 & -7.5\% &  &  &  &  \\
\hline\hlineType & \multicolumn{3}{l|}{Details \& Parameter} & Type
& \multicolumn{3}{l|}{ Details \& Parameter} \\ \hline1 &
\multicolumn{3}{l|}{Additive offset (\%MFD)} & 4 &
\multicolumn{3}{l|}{ Additive drift (\%MFD)} \\ \hline 2 &
\multicolumn{3}{l|}{Additive incipient offset (\%MFD)} & 5 &
\multicolumn{3}{l|}{Abrupt freezing (-)} \\ \hline 3 &
\multicolumn{3}{l|}{Noise (variance \%MFD)} & 6 &
\multicolumn{3}{l|}{ Multiplicative offset (divided by)} \\ \hline
\end{tabular}}
\end{table}

\begin{figure}[tbph]
\centering \subfigure{
   \includegraphics[width = 2in]{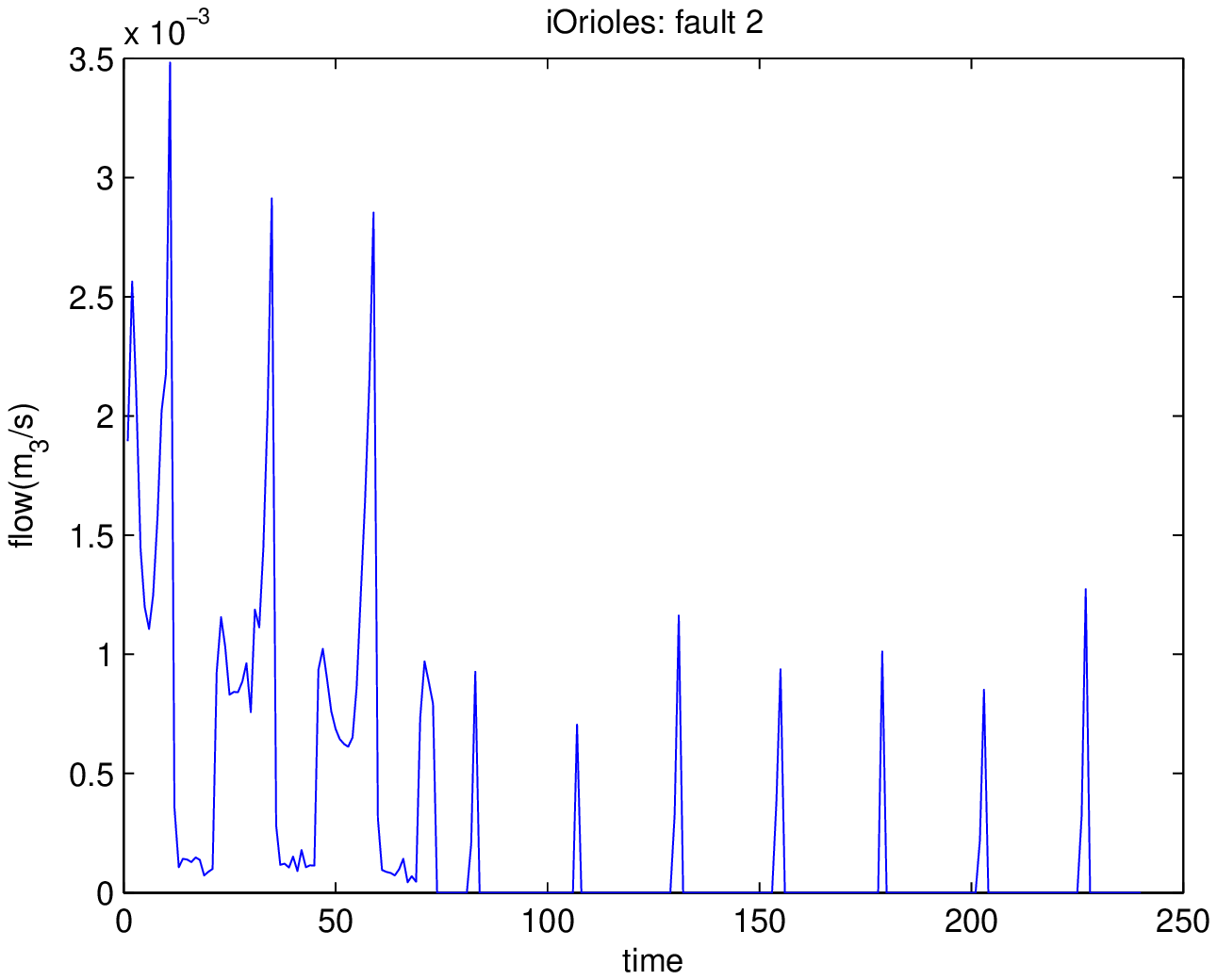}} \hspace{0.01in}
\subfigure{
    \includegraphics[width = 2in]{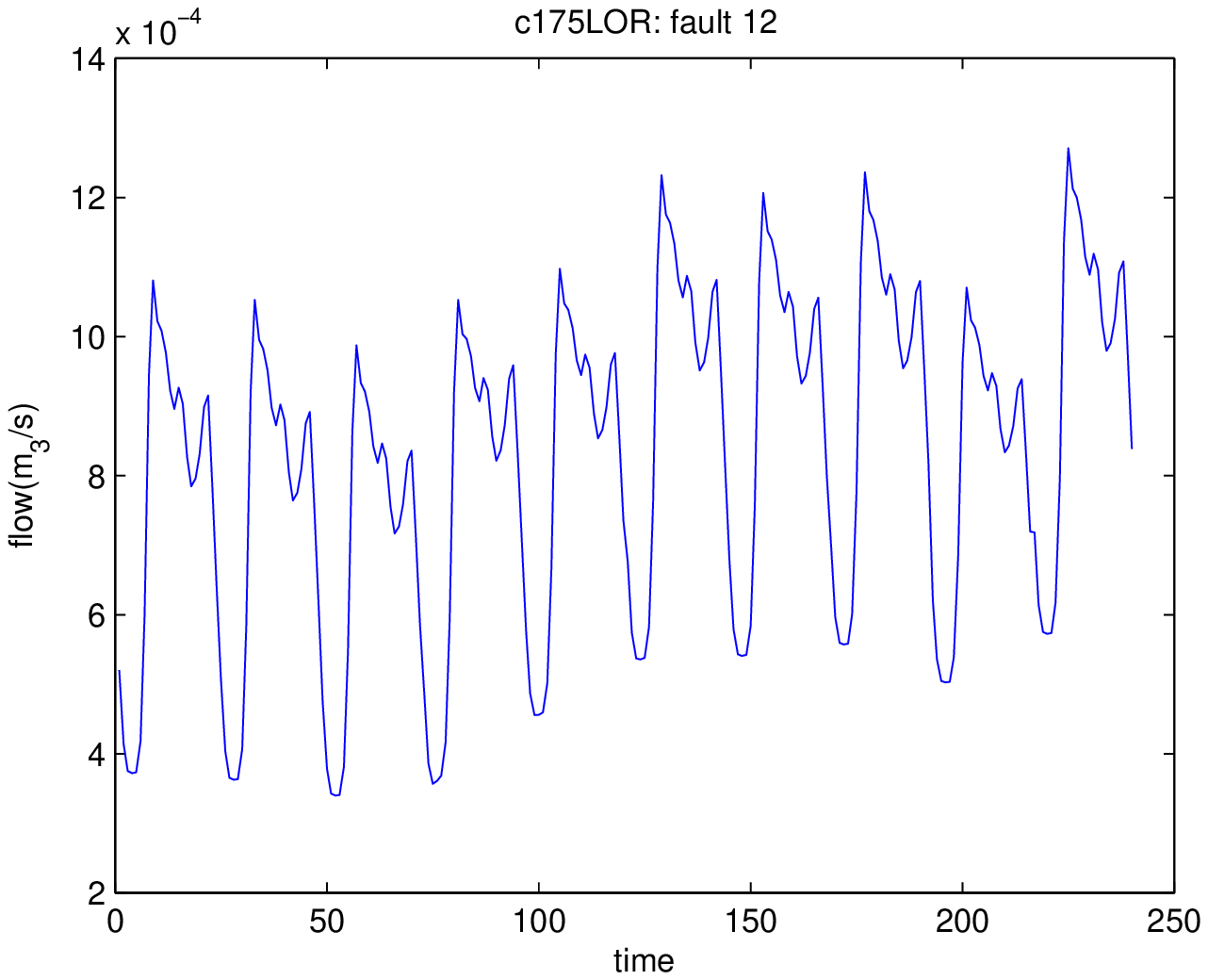}} \hspace{0.01in}
\subfigure{
    \includegraphics[width = 2in]{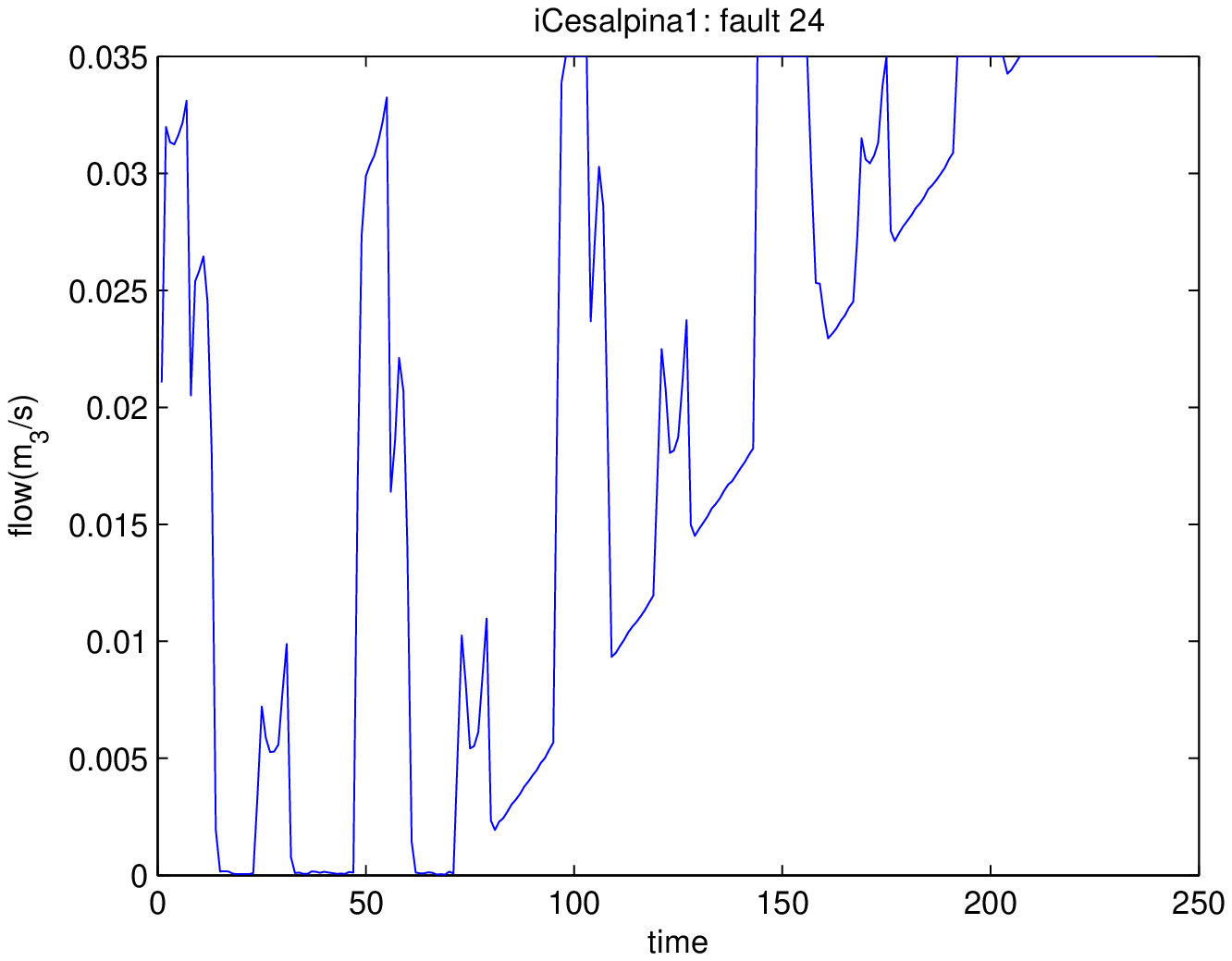}} \hspace{0.01in}
\subfigure{
    \includegraphics[width = 2in]{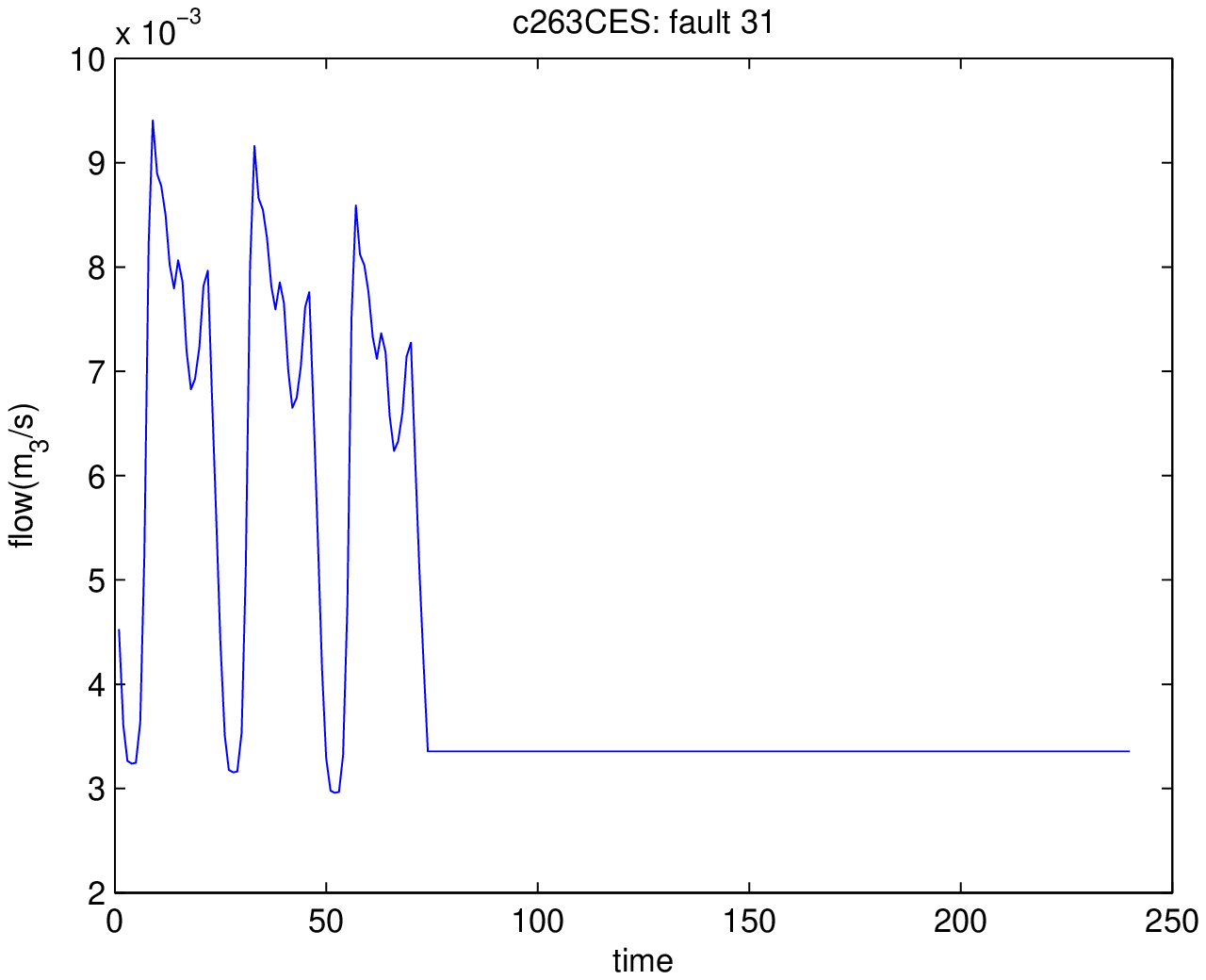}}
\caption{Examples of Faulty Signals}
\label{fault}
\end{figure}
As there are two subsystems, two deterministic reservoir computing models,
each with 25 nodes in the reservoir, have been employed in the proposed
framework.

\subsection{Comparisons and Evaluations}

\begin{table*}[tbp]
\caption{Comparisons of model space based approach and signal based approach
using supervised learning techniques. The reported results are based on 10
runs of 5-fold cross validation.}
\label{verse}\centering%
\resizebox {5in }{!}{
\begin{tabular}{|c|c|c|c|c|c|c|c|c|}
\hline
Algorithm & \multicolumn{2}{|c}{NARMA} & \multicolumn{2}{|c|}{Van der Pol} &
\multicolumn{2}{|c|}{Three Tank} & \multicolumn{2}{|c|}{Water} \\ \hline
& Model & Signal & Model & Signal & Model & Signal & Model & Signal \\
\hline CART & \textbf{0.00(0.00)} & 0.33(0.01) & \textbf{0.07(0.01)}
& 0.11(0.01) & \textbf{0.01(0.00)} & 0.02(0.00) &
\textbf{0.06(0.01)} & 0.11(0.00) \\ \hline SVMs &
\textbf{0.00(0.00)} & 0.07(0.01) & \textbf{0.05(0.01)} & 0.07(0.01)
& \textbf{0.00(0.00)} & \textbf{0.00(0.00)} & \textbf{0.06(0.00)} &
0.14(0.00)
\\ \hline
OCS & \textbf{0.04(0.01)} & 0.32(0.01) & \textbf{0.15(0.01)} &
0.27(0.01)
& \textbf{0.02(0.01)} & 0.10(0.01) & \textbf{0.09(0.01)} & 0.23(0.00) \\
\hline
Bagging & \textbf{0.00(0.00)} & 0.24(0.01) & \textbf{0.01(0.00)} & 0.07(0.00)
& \textbf{0.00(0.00)} & 0.01(0.01) & \textbf{0.04(0.01)} & 0.08(0.01) \\
\hline
Boosting & \textbf{0.00(0.00)} & 0.33(0.01) & \textbf{0.15(0.01)} &
0.22(0.01) & \textbf{0.01(0.00)} & 0.04(0.00) & \textbf{0.07(0.01)} &
0.16(0.00) \\ \hline
\end{tabular}}
\end{table*}

\begin{table*}[tbp]
\caption{Comparisons of several algorithms in terms of fault detection
ability, i.e. fault detection rate (FDR) and false alarm rate (FAR).}
\label{faultDetection}\centering%
\resizebox {5in }{!}{
\begin{tabular}{|c|c|c|c|c|c|c|c|c|}
\hline
& \multicolumn{2}{|c}{NARMA} & \multicolumn{2}{|c|}{Van der Pol} &
\multicolumn{2}{|c|}{Three Tank} & \multicolumn{2}{|c|}{Barcelona Water} \\
\hline
Algorithm & FDR & FAR & FDR & FAR & FDR & FAR & FDR & FAR \\ \hline
T2 & 0.9072 & 0.1000 & 0.3009 & 0.0998 & 0.2311 & 0.0999 & 0.2316 & 0.1384
\\ \hline
DBscan & 1 & 0.0917 & 0.9146 & 0.2317 & 0.8958 & 0.0683 & 0.7981 & 0.1368 \\
\hline OCS-Model & 1 & 0.1102 & 0.9310 & 0.0509 & 0.8521 & 0.1082 &
0.9313 & 0.2683 \\ \hline OCS-Signal & 0.7042 & 0.2097 & 0.7686 &
0.2104 & 0.7521 & 0.2082 & 0.4920 & 0.3796 \\ \hline AP-Model &
\textbf{1} & \textbf{0} & 1.0000 & 0.3405 & 0.8407 & 0.1128 & 0.9014
& 0.2678 \\ \hline
AP-Signal & 1 & 0.5427 & 1.0000 & 0.7405 & 0.7155 & 0.2387 & 0.8879 & 0.2458 \\
\hline
ARMAX-OCS & 0.9882 & 0.0517 & 0.8727 & 0 & 0.9776 & 0 & 0.7369 & 0.1588 \\
\hline
RC-OCS & 0.9747 & 0.0558 & 0.9762 & 0.0158 & 0.8387 & 0 & 0.8271 & 0.1079 \\
\hline DRC-OCS(Sampling) & 0.9789 & 0 & 0.9804 & 0 & \textbf{0.9926}
& \textbf{0} & 0.9327 & 0.0817 \\ \hline
DRC-OCS & 0.9921 & 0 & \textbf{0.9818} & \textbf{0} & 0.9919 & 0 & \textbf{0.9762} & \textbf{0.0473} \\ \hline
\end{tabular}}
\end{table*}

\begin{table*}[tbp]
\caption{Comparisons of several algorithms in terms of fault isolation
ability.}
\label{faultIsolation}\centering%
\resizebox{5in }{!}{
\begin{tabular}{|c|c|c|c|c|c|c|c|c|}
\hline
& \multicolumn{4}{|c}{NARMA (3 classes)} & \multicolumn{4}{|c|}{Van der Pol
(4 classes)} \\ \hline
Algorithm & Classes & Precision & Recall & Specificity & Classes & Precision
& Recall & Specificity \\ \hline
DBscan & 4 & 0.6690 & 0.7650 & 0.8825 & 10 & 0.7629 & 0.6842 & 0.8018 \\
\hline
AP-Model & 271 & 0.9699 & 0.9698 & 0.9899 & 367 & 0.8778 & 0.8757 & 0.9585
\\ \hline
ARMAX-OCS & 5 & 0.9354 & 0.9229 & 0.9615 & 2 & 0.4309 & 0.4880 & 0.7868 \\
\hlineRC-OCS & 3 & 0.9637 & 0.9615 & 0.9808 & 6 & 0.9606 & 0.9583 & 0.9861 \\
\hline DRC-OCS(Sampling) & 3 & 0.9683 & 0.9692 & 0.9914 & 5 & 0.9617
& 0.9726 & 0.9819 \\ \hline DRC-OCS & 3 & \textbf{0.9861} &
\textbf{0.9858} & \textbf{0.9929} & 5 & \textbf{0.9736} &
\textbf{0.9731} & \textbf{0.9910} \\ \hline &
\multicolumn{4}{|c}{Three Tank (4 classes)} & \multicolumn{4}{|c|}{
Barcelona Water (32 classes)} \\ \hline Algorithm & Classes &
Precision & Recall & Specificity & Classes & Precision & Recall &
Specificity \\ \hline
DBscan & 14 & 0.8742 & 0.7561 & 0.9253 & 61 & 0.8019 & 0.7326 & 0.8654 \\
\hline
AP-Model & 272 & 0.9713 & 0.9704 & 0.9901 & 654 & 0.9366 & 0.9428 & 0.9751
\\ \hline
ARMAX-OCS & 5 & 0.9914 & 0.9923 & 0.9984 & 57 & 0.7826 & 0.7419 & 0.8237 \\
\hline
RC-OCS & 9 & 0.9182 & 0.8788 & 0.9596 & 44 & 0.8913 & 0.8942 & 0.9263 \\
\hline DRC-OCS(Sampling) & 7 & \textbf{0.9940} & \textbf{0.9949} &
\textbf{0.9988} & 39 & 0.9219 & 0.9310 & 0.9513 \\ \hline
DRC-OCS & 10 & 0.9931 & 0.9931 & 0.9977 & 48 & \textbf{0.9538} & \textbf{0.9640} & \textbf{0.9871} \\ \hline
\end{tabular}}
\end{table*}

This section will first report the comparisons of several supervised
algorithms applied in the model space and signal space, respectively, and
then evaluate those algorithms listed in Table \ref{AlgoPara} in terms of
fault detectability and fault isolationability.

In above section, the model space and signal space have been illustrated by
the MDS algorithm. However, due to the high dimensionality, the
visualizations might not reveal the real relationship of these data points
in the high dimensional space. In order to compare the model space and
signal space based approaches, Table \ref{verse} reports the comparisons of
the representations of model space and signal space using a number of
supervised learning algorithms, including classification and regression
trees (CART), support vector machines (SVMs), one class support vector
machine (OCS), Bagging (100 trees) and Adaboosting (100 trees).

In the signal space approach, the order $p$ will be selected in the range $%
[1,30]$ by 5-fold cross validation approach. The parameters of SVMs and
one-class SVMs are optimized by 5-fold cross validation. The parameters in
CART, Bagging and Adaboosting follow the defaults in MATLAB.

The reported results in Table \ref{verse} are based on 10 runs of 5-fold
cross validation. In Table \ref{verse}, model space representation usually
achieves lower error rate. In some cases, e.g. CART/SVMs in NARMA and
SVM/Bagging in three tank system, model space representation can even
achieve 100\% accuracy. These results are consistent with those MDS
visualizations, and confirm the benefits to use model space rather than
signal space in fault diagnosis.

In fault diagnosis, the first step is to discriminate faults from normal
situations. Table \ref{faultDetection} reports fault detection results using
a number of algorithms listed in Table \ref{AlgoPara}. The parameters
related to DBscan, one-class SVM and ARMAX are optimized by 5-fold cross
validation in the normal period. In this table, fault detection rate (FDR)
and false alarm rate (FAR) are employed as two metrics.

According to Table \ref{faultDetection}, model space based algorithms, such
as DRC-OCS, RC-OCS, are superior to other algorithms. Since deterministic
reservoir is more stable than random reservoir and there is no model
assumption in DRC\footnote{%
ARMAX model assumes the model order and ARMAX-OCS might not perform well on
signals with incorrect model assumption.}, DRC-OCS is better than RC-OCS and
ARMAX-OCS.

Although the sampling method of DRC-OCS could potentially obtain better
estimates when the readout parameters are non-uniform, it would require
dense sampling points, i.e. large window size $m$ in this case, with
increased computational cost. However, due to real-time requirements and
computational restrictions, the windows size should be restricted for prompt
response to faults. Hence, DRC-OCS (sampling) is often inferior to DRC-OCS.

The statistical-test based algorithm T2 acts are a base line algorithm and
it usually has a lower FDR and a fair FAR. DBscan and affinity propagation
(AP) are clustering based algorithms. As these clustering algorithms do not
make use of the information that the first $t$ steps are normal, these
algorithms did not perform well in the four applications.

In time-varying environment, there may be unanticipated fault scenarios that
haven't been encountered before. In this paper, we proposed a dynamic fault
library construction framework and its application on fault isolation. These
results are reported in Table \ref{faultIsolation}.

In Table \ref{faultIsolation}, we first report the true number of classes
and the discovered classes (i.e. number of faults plus normal class) using a
number of algorithms for each data set\footnote{%
Due to the assumption that the type of faults are unknown in advance, these
compared algorithms always discover more faults than true number of faults
by decomposing each true fault to a number of small fault segments.}. Then,
we report the fault isolation performance of these algorithms in terms of
precision, recall and specificity.

Since the number of discovered faults does not equal to the true number of
faults, we compare each true cluster $\Lambda_{i}$ and these discovered
clusters and merge those clusters with maximizing overlap with $\Lambda_{i}$
to a pseudo-cluster $\tilde{\Lambda}_{i}$. The performance metrics are
obtained by comparing $\Lambda$ and $\tilde{\Lambda}$.

Based on Table \ref{faultIsolation}, DRC-OCS usually outperforms other
algorithms under these three metrics. AP-model performs well on the
isolation stage, but it often generates too many sub-faults in the library,
e.g. 270 sub-faults verse 2 faults.

In the three ``learning in the model space'' approaches, i.e. DRC-OCS,
RC-OCS and ARMAX-OCS, DRC-OCS is the best and ARMAX-OCS is the most inferior
one as it requires the model order selection for different applications.
Without prior information for complex applications, it is usually difficult
to select the model order. With limited sampling points due to real-time
requirement, the sampling method of DRC-OCS is often inferior to DRC-OCS,
though it often outperforms other approaches.

Based on the results presented in Table \ref{verse}, \ref{faultDetection}
and \ref{faultIsolation}, the proposed approach DRC-OCS achieves the best
results and these results also confirmed that \textquotedblleft learning in
the model space\textquotedblright\ is an effective framework for fault
diagnosis.

\section{Conclusion}

\label{conclusion}

In this paper, an effective cognitive fault diagnosis framework has been
proposed to tackle the challenges in complex engineering systems in
time-varying or un-formulated environment. Instead of investigating the
fault diagnosis in the signal space, this paper introduces ``learning in the
model space'' framework that represents the multiple-input and
multiple-output data as a series of models fitted using a rolling window. By
investigating the characteristic of these fitted models using learning
approach in the model space, we can identify and isolate faults effectively,
and dynamically construct a fault library.

This contribution applies deterministic reservoir models to fit the MIMO
data, since reservoir models are generic to fit a wide variety of dynamical
features of the input driven signals, and the deterministic reservoir models
further simplify the model structure and thus improve the fitting
performance.

To rigorously investigate these fitted models for fault diagnosis, this
paper demonstrates the application of the distance definition in the model
space for linear readout models. The model distance differs from the squared
Euclidean distance of the readout parameters, indicating that more
importance is given to the `offset' than `orientation' of the readout
mapping. We also present the estimated forms of model distance by using
either sampling methods or a Gaussian mixture model when the domain of
readout-parameters is non-uniform.

By replacing the data distance matrix with the \emph{model distance} matrix,
one-class SVMs are able to ``learn'' in the model space to identify
normal/abnormal models. To accommodate unknown faults, the algorithm
``incremental one class learning in the model space'' is proposed to
identify and isolate faults, and simultaneously construct the fault library.

To evaluate this proposed framework with other related fault diagnosis
approaches, three benchmark systems and one simulated model for Barcelona
water system have been employed. The results confirm both the benefits to
represent MIMO data in the model space and the effectiveness of ``learning
in model space'' framework.

``Learning in the model space'' is an effective framework for complex data
representation and fault diagnosis. Instead of using reservoir models and
one class SVMs as fitting and discriminating models, respectively, there
should be other effective opinions or combinations for various application
systems, which consist of our future work.

\section*{Acknowledgment}

This work is supported by the European Union Seventh Framework Programme
under grant agreement No. INSFO-ICT-270428. This work has benefitted from
many discussions with the members of the iSense project team.

\bibliographystyle{IEEEtran}
\bibliography{IEEEabrv,bibtex}

\end{document}